\documentclass[10pt,twocolumn,letterpaper]{article}

\usepackage{style}
\PassOptionsToPackage{numbers, compress,sort}{natbib}

\usepackage{bm}  

\usepackage[utf8]{inputenc} 
\usepackage[T1]{fontenc}    
\usepackage{url}            
\usepackage{booktabs}       
\usepackage{amsfonts}       
\usepackage{nicefrac}       
\usepackage{microtype}      
\usepackage{graphicx}
\usepackage{subcaption}
\usepackage{booktabs} 
\usepackage{natbib}

\usepackage{here}
\usepackage{graphicx}
\usepackage{caption}
\usepackage[ruled,noend]{algorithm2e}
\usepackage{amsmath,amssymb,amsfonts,amsbsy,amsfonts,latexsym}
\usepackage{amsthm}
\usepackage{multirow}
\usepackage{makecell}
\usepackage[labelfont=bf,textfont=it,belowskip=0pt,aboveskip=5pt,tableposition=top]{caption}
\usepackage{xcolor}
\usepackage{colortbl}

\usepackage{wrapfig}

\definecolor{colorA}{RGB}{189,201,225}
\definecolor{colorB}{RGB}{103,169,207}
\definecolor{colorC}{RGB}{ 28,144,153}
\definecolor{colorD}{RGB}{  1,108, 89}

\newcolumntype{R}{>{\columncolor{gray!40}}r}
\newcolumntype{L}{>{\columncolor{gray!40}}l}
\newcolumntype{C}{>{\columncolor{gray!40}}c}

\usepackage{tabularx,colortbl,xcolor}
\usepackage{multirow}
\usepackage[normalem]{ulem}
\useunder{\uline}{\ul}{}

\newcommand\Ga{}

\newcommand\Gc{\rowcolor{gray!30}}

\usepackage{enumitem}

\usepackage{xparse}

\graphicspath{{figs}}
\DeclareGraphicsExtensions{.pdf,.png}
\usepackage{pgfplots, pgfplotstable}
\usepackage[colorlinks=true,
filecolor=black,
citecolor=blue,
linkcolor=blue,
urlcolor=blue]{hyperref}

\DeclareMathOperator{\RR}{\mathbb{R}}
\DeclareMathOperator*{\argmax}{arg\,max}

\newcommand{\M}{\mathcal M}

\def\g{{\bf g}}

\def\H{{\bf H}}

\def\x{{\bf x}}

\def\y{{\bf y}}
\def\Z{{\bf Z}}
\def\z{{\bf z}}
\def\0{{\bf 0}}
\def\1{{\bf 1}}

\def\M{\mathcal M}

\NewDocumentCommand{\var}{O{s} m O{}}{%
  \ensuremath{#1_{#2}^{#3}}
}
\usepackage{siunitx}

\DeclareMathOperator*{\argmin}{arg\,min}

\definecolor{light-gray}{gray}{0.80}

\renewcommand\paragraph{\subsubsection*}

\newcommand\btimes{{\bm{\times}}}

\newcommand\fref{Fig.~\ref}

\newcommand\tref{Table~\ref}
\SetKwInOut{Parameter}{parameter}
\newcommand\cmt[1]{\tcp*[r]{\scriptsize \color{gray!80!black}#1}}

\newcommand\eps{ \epsilon}

\def\x{{\bf x}}
\def\H{{\bf H}}
\def\g{{\bf g}}

\def\0{{\bf 0}}
\def\s{{\bf s}}

\newcommand\atsign{@}

\graphicspath{{figures/}}

\stylefinalcopy 
\ifstylefinal\fi
\begin{document}
\title{Trust Region Based Adversarial Attack on Neural Networks}
\author{Zhewei Yao$^1$~~~Amir Gholami$^1$~~~Peng Xu$^2$~~~Kurt Keutzer$^1$~~~Michael W. Mahoney$^1$\\
$^1$University of California, Berkeley~~~$^2$ Stanford Univeristy\\
$^1$\{zheweiy, amirgh, keutzer, and mahoneymw\}\atsign berkeley.edu,
$^2$pengxu\atsign stanford.edu
}

\maketitle
\begin{abstract}
Deep Neural Networks are quite vulnerable to adversarial perturbations. 
Current state-of-the-art adversarial attack methods typically require very time consuming hyper-parameter tuning,
or require many iterations to solve an optimization based adversarial attack.
To address this problem, we present a new family of trust region based adversarial attacks, with the goal of computing adversarial perturbations efficiently. 
We propose several attacks based on variants of the trust region optimization method.
We test the proposed methods on Cifar-10 and ImageNet datasets using several different models including AlexNet, ResNet-50, VGG-16, and DenseNet-121 models.
Our methods achieve comparable results with the Carlini-Wagner (CW) attack, but with significant speed up of up to $37\times$, for the VGG-16 model on a Titan Xp GPU.
For the case of ResNet-50 on ImageNet, we can bring down its classification accuracy to less than 0.1\% with at most $1.5\%$ relative $L_\infty$ (or $L_2$) perturbation requiring
only $1.02$ seconds as compared to $27.04$ seconds for the CW attack.
We have open sourced our method which can be accessed at~\cite{TRcode}.
\end{abstract}

\section{Introduction}\label{sec:intro}

\begin{figure}
\begin{center}
  \includegraphics[width=.47\textwidth]{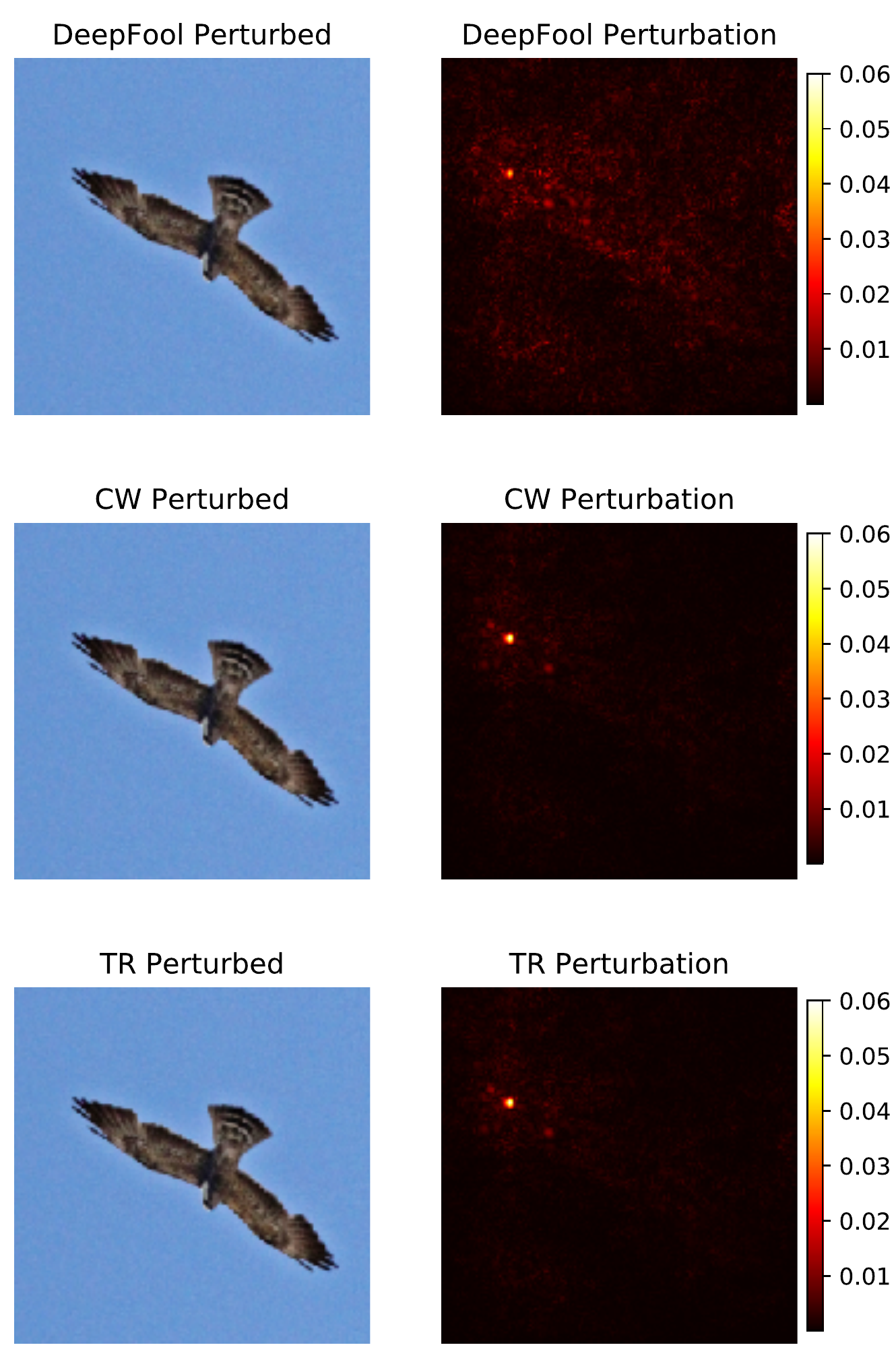} 
\end{center}
\caption{
An example of DeepFool, CW, and our TR attack on AlexNet, with $L_2$ norm. 
Both CW and TR perturbations are smaller in magnitude and more targeted than DeepFool's ($2\times$ smaller here). 
TR attack obtains similar perturbation as CW, but $\bf 15\btimes$ faster.
In the case of the VGG-16 network, we achieve an even higher speedup of $\bf 
37.5\btimes$ (please see~\fref{f:efficient} for timings).
}
\label{f:adv_example1}
\end{figure}

Deep Neural Networks (DNNs) have achieved impressive results in many research areas, such as classification,
object detection, and natural language processing. 
However, recent studies have shown that DNNs are often not robust to adversarial perturbation of the input data~\cite{szegedy2013intriguing,fgsm}.
This has become a major challenge for DNN deployment, and significant research has been performed to address this. 
These efforts can be broadly classified in three categories:
(i) research into finding strategies to defend against adversarial inputs (which has so far been largely unsuccessful);
(ii) new attack methods that are stronger and can break the proposed defense mechanisms; and 
(iii) using attack methods as form of implicit adversarial regularization for training neural networks~\cite{yao2018large,robust,yao2018hessian}.
Our interest here is mainly focused on finding more effective attack methods that could be used in the latter two directions. 
Such adversarial attack methods can be broadly classified into two categories: 
white-box attacks, where the model architecture is known; and 
black-box attacks, where the adversary can only perform a finite number of queries and observe the model behaviour.
In practice, white-box attacks are often not feasible, but recent work has shown that some adversarial attacks
can actually transfer from one model to the other~\cite{moosavi2017universal}. 
Therefore, precise knowledge of the target DNN  may actually not be essential. 
Another important finding in this direction is the existence of an \textit{adversarial patch}, i.e., a small set of pixels which, if added to an image, can fool the network. 
This has raised  important security concerns for applications such as autonomous driving, where addition of such an adversarial patch to traffic signs could fool the~system~\cite{brown2017adversarial}.

Relatedly, finding more \textit{efficient} attack methods is important for evaluating defense strategies, and this is the main focus of our paper. 
For instance, the seminal work of~\cite{carlini2017towards} introduced a new type of optimization based attack, commonly referred to as CW (Carlini-Wagner) attack, which illustrated that defensive distillation~\cite{papernot2016distillation} can be broken with its stronger attack.
The latter approach had shown significant robustness to the fast gradient sign attack method~\cite{fgsm}, but not when tested against the stronger CW attack.
Three metrics for an \textit{efficient attack} are the speed with which such a perturbation can be computed,
the magnitude or norm of the perturbation that needs to be added to the input to fool the network, and the transferability of the attack to other networks. 
Ideally (from the perspective of the attacker), a stronger attack with a smaller perturbation magnitude is desired, so that it could be undetectable (e.g. an adversarial patch that is harder to detect by~humans).

\begin{figure*}[!htbp]
\begin{center}
  \includegraphics[width=.9\textwidth]{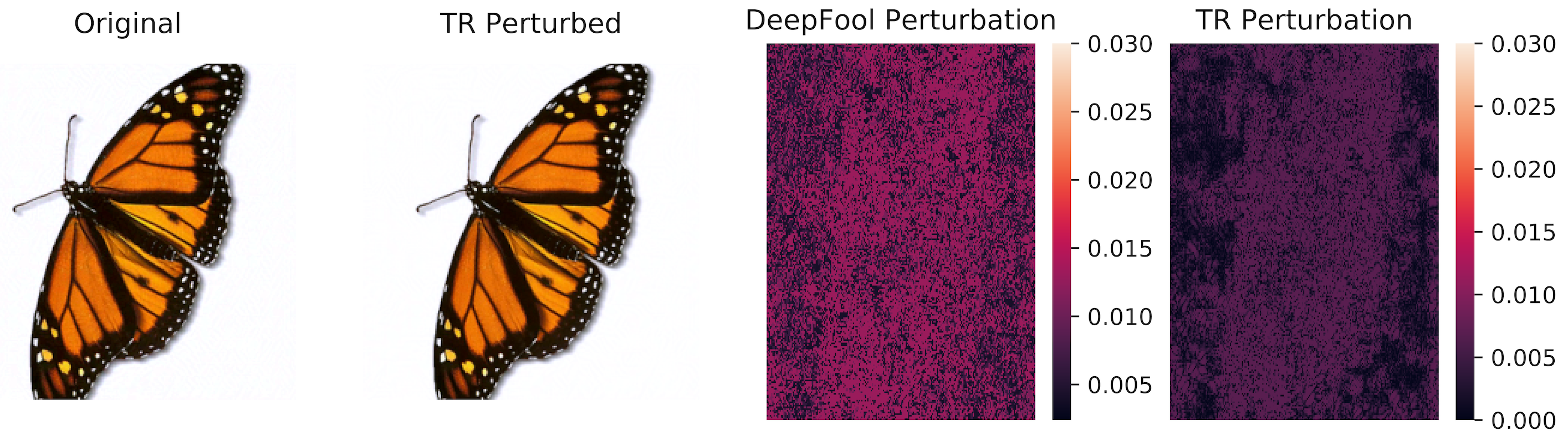}
\end{center}
\caption{
An example of DeepFool and TR attack on VGG-16, with $L_\infty$ norm. 
The first pattern is the original image. 
The second pattern is the image after TR attack. 
The final two patterns are the perturbations generated by DeepFool and TR. 
The TR perturbation is smaller than DeepFool's ($1.9\times$ smaller). 
Also, the TR perturbation is more concentrated around the butterfly.
}
\label{f:adv_example2}
\end{figure*}

In this work, we propose a novel trust region based attack method.
Introduced in~\cite{steihaug1983conjugate}, 
trust region (TR) methods form a family of numerical optimization methods for solving non-convex optimization problems~\cite{NW06}.
The basic TR optimization method works by first defining a region, commonly referred to as \textit{trust region}, around the current point in the optimization landscape, in which a (quadratic) model approximation is used to find a descent/ascent direction.
The idea of using this confined region is due to the model approximation error.
In particular, the trust region method is designed to improve upon vanilla first-order and second-order methods, especially in the presence of non-convexity.

We first consider a first-order TR method, which uses gradient information for attacking the target DNN model and adaptively adjusts the trust region. 
The main advantage of first-order attacks is their computational efficiency and ease of implementation.
We show that our first-order TR method significantly reduces the over-estimation problem (i.e. requiring very large perturbation to fool the network),
resulting in up to $3.9\times$ reduction in the perturbation magnitude, as compared to DeepFool~\cite{moosavi2016deepfool}.
Furthermore, we show TR is significantly faster than the CW attack (up to $37\times$), while achieving similar attack performance.
We then propose an adaptive TR method, where we adaptively choose the TR radius based on the model approximation to further speed up the
attack process.
Finally, we present the formulation for how our basic TR method could be extended to a second-order TR method, which could be useful for cases with significant non-linear decision boundaries, e.g., CNNs with Swish activation function~\cite{ramachandran2018searching}. 
In more detail, our main contributions are the following:

\begin{itemize}
\item 
We cast the adversarial attack problem into the optimization framework of TR methods.
This enables several new attack schemes, which are easy to implement and are significantly more
effective than existing attack methods (up to $3.9\times$, when compared to DeepFool). 
Our method requires a similar perturbation magnitude, as compared to CW, but it can compute the perturbation significantly faster (up to $37\times$), as it does not require 
extensive hyper-parameter tuning.
\item  Our TR-based attack methods can adaptively choose the perturbation magnitude in every iteration.
This removes the need for expensive hyper-parameter tuning, which is a major issue with the existing optimization based methods.
\item 
Our method can easily be extended to second-order TR attacks, which could be useful for non-linear activation functions.
With fewer iterations, our second-order attack method outperforms the first-order attack method.
\end{itemize}

\textbf{Limitations.}
We believe that it is important for every work to state its limitations (in general, but in particular in this area).
We paid special attention to repeat the experiments multiple times, and we considered multiple different DNNs on different datasets to make sure the results are general. 
One important limitation of our approach is that a na\"{\i}ve implementation of our second-order method requires computation of Hessian matvec backpropogation, which is very expensive for DNNs.  Although the second-order TR attack achieves better results, as compared to the first-order TR attack, this additional computational cost could limit its usefulness in certain applications.
Moreover, our method achieves similar results as CW attack, but significantly faster. However, if we ignore the strength of the attack, then the DeepFool attack
is faster than our method (and CW's for that matter).
Although such comparison may not be fair, as our attack is stronger.
However, this may be an important point for certain applications
where maximum speed is needed.

\section{Background}
In this section, we review related work on adversarial attacks. 
Consider $\x\in\RR^n$ as input image, and $\y\in\RR^c$ the corresponding label.
Suppose $\M(\x;\theta)=\hat \y$ is the DNN's prediction probabilities, with $\theta$ the model parameters and $\hat{\y}\in\RR^c$ the vector of probabilities.
We denote the loss function of a DNN as $\mathcal{L}(\x,\theta,\y)$. 
Then, an adversarial attack is aimed to find a (small) perturbation, $\Delta \x$, such that:
\begin{equation*}
    \argmax(\M(\x+\Delta \x;\theta))=\argmax(\hat \y) \not= \argmax(\y).
\end{equation*}

There is no closed form solution to analytically compute such a perturbation. However, several different
approaches have been proposed by solving auxiliary optimization or analytical approximations to solve
for the perturbation. For instance,
the Fast Gradient Sign Method (FGSM)~\cite{fgsm} is a simple
adversarial attack scheme that works by directly maximizing the loss function $\mathcal{L}(\x,\theta,\y)$.
It was subsequently extended to an iterative FGSM~\cite{fgsm10}, which performs multiple gradient ascend steps to compute the adversarial perturbation, and is often more effective than FGSM in attacking the network. 
Another interesting work in this direction us DeepFool, which uses an approximate analytical
method.
DeepFool assumes that the neural network behave as an affine multiclass classifier, which allows
one to find a closed form solution.
DeepFool is based on ``projecting'' perturbed inputs to cross the decision boundary (schematically shown in~\fref{f:decision_boundary}), so that its classification is changed, and this was shown to outperform FGSM.
However, the landscape of the decision boundary of a neural network is not linear. 
This is the case even for ReLU networks with the softmax layer.
Even before the softmax layer, the landscape is \emph{piece-wise linear}, but this cannot be approximated
with a simple affine transformation.
Therefore, if we use the local information, we can overestimate/underestimate the adversarial perturbation needed to fool
the network.

The seminal work of~\cite{carlini2017towards} introduced the so-called CW attack, a more sophisticated way to directly solve for the $\Delta \x$ perturbation.
Here, the problem is cast as an optimization problem, where we seek to minimize the distance between the original
image and the perturbed one, subject to the constraint that the perturbed input
would be misclassified by the neural network. 
This work also clearly showed that defensive distillation, which at the time was believed to be a robust method to defend against adversaries, is not robust to stronger attacks.
One major disadvantage of the CW attack is that it is very sensitive to hyper-parameter tuning.
This is an important problem in applications where speed is important, as finding a good/optimal adversarial perturbation for a given input is very time consuming.
Addressing this issue, without sacrificing attack strength, is a goal of our work.

On another direction, adversarial training has been used as a defense method against adversarial attacks~\cite{robust}.
In particular, by using adversarial examples during training, one can obtain models that are more robust to attacks (but still not foolproof). 
This adversarial training was further extended to ensemble adversarial training~\cite{tramer2017ensemble},
with the goal of making the model more robust to black box attacks.
Other approaches have also been proposed to detect/defend against adversarial attacks~\cite{papernot2016distillation,metzen2017detecting}.
However, it has recently been shown that, with a stronger attack method, defense schemes such as distillation or obfuscated gradients can be broken~\cite{carlini2017adversarial, carlini2017towards,athalye2018obfuscated}.

A final important application of adversarial attacks is to train neural networks to obtain improved generalization, even in non-adversarial environments.
Multiple recent works have shown that adversarial training (specifically, training with mixed adversarial and clean data) can be used to train a neural network from scratch in order to achieve a better final \emph{generalization} performance~\cite{robust,sankaranarayanan2017regularizing,yao2018large,yao2018hessian}.
In particular, the work of~\cite{yao2018hessian} empirically showed that using adversarial training would result in finding
areas with ``flatter'' curvature.
This property has recently been argued to be an important parameter for generalization performance~\cite{keskar2016large}. 
Here, the speed with which adversarial perturbations can be computed is very important since it appears in the inner loop of the training process, and the training needs to be performed for many epochs.

\section{Trust Region Adversarial Attack}\label{sec:method}
Let us denote the output of the DNN before the softmax function to be $\z=\Z(\x;\theta)\in\RR^c$. Therefore, we will have:
\[
\M(\x;\theta)=softmax\big(\Z(\x; \theta)\big)=\hat \y.
\]

\noindent Denote $\y_t$ to be the true label of $\x$, and $\z_t = \arg\max \z$ to be the prediction output of $\M(\x;\theta)$.
For clarification, note that $\z_t$ is only the same as $\y_t$ if the
neural network makes the correct classification.
An adversarial attack seeks to find  $\Delta \x$ that fools the DNN, that is:
\begin{equation}\label{eqn:basic_prob}
\arg\min_{\|\Delta \x\|_p} \arg\max \Z(\x+\Delta\x; \theta) \not=\y_t,
\end{equation}

\begin{figure}
\centering
\includegraphics[width=0.30\textwidth]{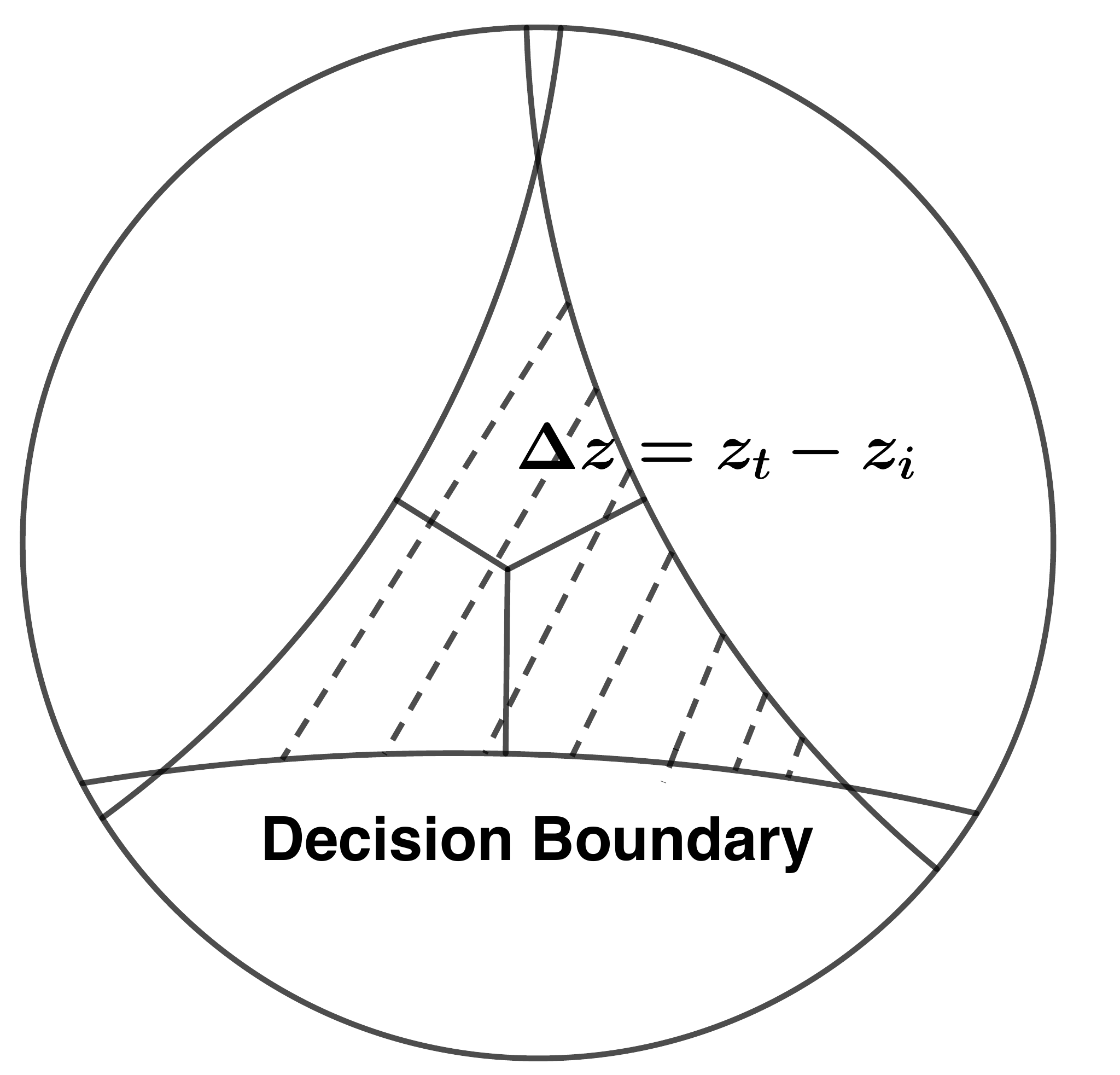}
\caption{Schematic illustration of the decision boundaries for a classification problem. Points mapped to the hashed region, are classified with the same label.
}
\label{f:decision_boundary}
\end{figure}

\begin{algorithm}[tbp]
\SetKwData{Left}{left}
\SetKwData{This}{this}
\SetKwData{Up}{up}
\SetKwFunction{Union}{Union}
\SetKwFunction{FindCompress}{FindCompress}
\SetKwInOut{Input}{input}
\SetKwInOut{Output}{output}
\Input{Image $\x^0$, label $\y$, initial radius $\eps^0$, threshold $\sigma_1$ and $\sigma_2$, radius adjustment rate $\eta$}
\Output{Adversarial Image}
\BlankLine
\emph{$\Delta\x=0$, $j=0$}\cmt{Initialization}
\emph{Using scheme to choose the attacking index $i$}\cmt{Index Selection}
\While {$\arg\max\Z(\x^j)=\arg\max \y$}{
    \emph{$\Delta\x_{tmp}=\arg\min_{\|\Delta\x^j\|\leq\eps^j} m^j(\Delta\x^j) =\arg\min_{\|\Delta\x^j\|\leq\eps^j} \langle\Delta\x^j, \g^j_{t,i}\rangle + \frac12\langle\Delta\x^j, \H^j_{t,i}\Delta\x^j\rangle$}
    \emph{$\x^{j+1}=clip(\x^j+\Delta\x_{tmp}, \min(\x),\max(\x))$}\cmt{Update}
    \emph{$\rho=\frac{(\z^{j+1}_t-\z^{j+1}_i)-(\z^{j+1}_t-\z^{j+1}_i)}{m^j(\Delta\x^j)}$}\cmt{Compute Ratio}
    \If {$\rho>\sigma_1$}{
    $\epsilon^{j+1}=\min\{\eta\epsilon^j,\epsilon_{\max}\}$
    }
    \ElseIf{$\rho<\sigma_2$}{
    $\epsilon^{j+1}=\min\{\epsilon^j/\eta, \epsilon_{\min}\}$
    }
    \Else{
    $\epsilon^{j+1}=\epsilon^j$
    }
    $j =j+ 1$
}

\caption{Trust Region Attack}
\label{alg:1}
\end{algorithm}

\begin{figure*}[!htbp]
\centering
\includegraphics[width=0.45\textwidth]{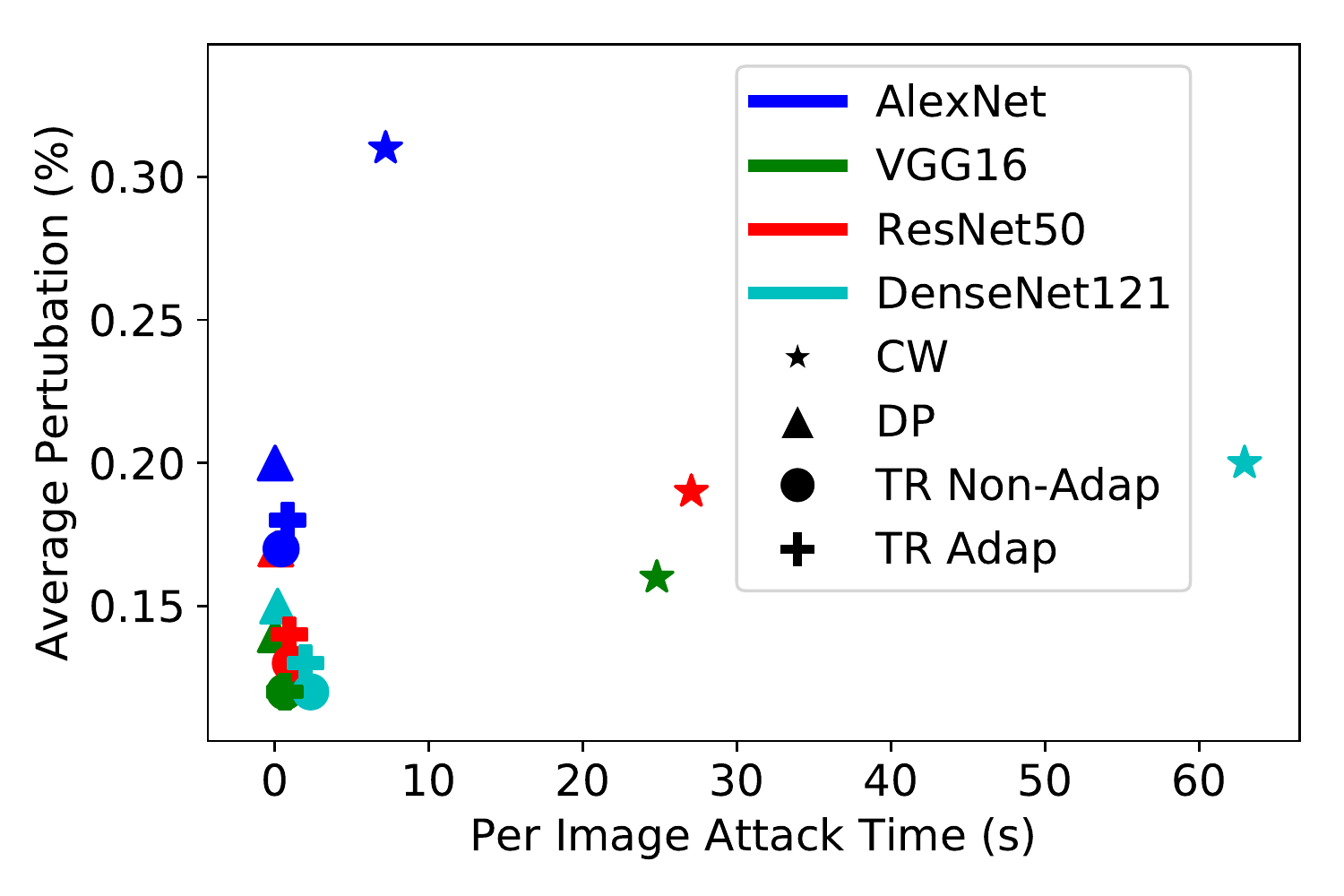}
\includegraphics[width=0.45\textwidth]{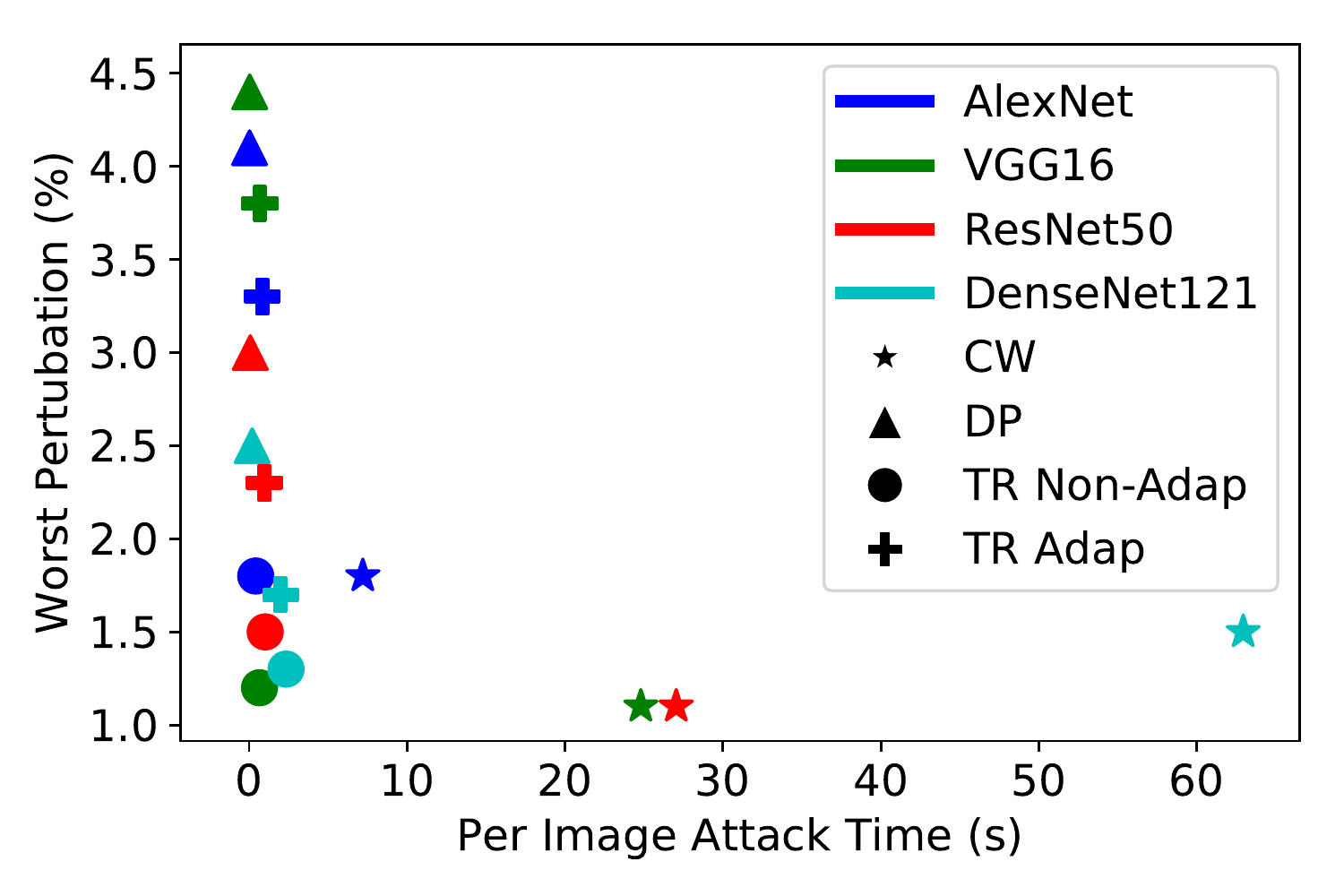}
\caption{The two subfigures show, for various neural networks, the time to compute the adversarial attack (x-axis) and the perturbation needed by that attack method to fool an image (y-axis), corresponding to ImageNet results in~\tref{tab:imagenet}.
On the left, the y-axis is plotted for average perturbation; and on the right, for the worst case perturbation.
An attack that achieves smaller perturbation in shorter time is preferred.
Different colors represent different models, and different markers illustrate the different attack methods. 
Observe that our TR and TR Adap methods achieve similar perturbations as CW but with significantly less time (up to $\bf 37.5\btimes$). 
}
\label{f:efficient}
\end{figure*}

\noindent  
where $\|\cdot\|_p$ denotes the $L_p$ norm of a vector.
It is often computationally infeasible to solve \eqref{eqn:basic_prob} exactly.
Therefore, a common approach is to approximately solve
\eqref{eqn:basic_prob}~\cite{carlini2017towards,fgsm,szegedy2013intriguing}.
To do so, the problem can be formulated as follows:
\begin{equation}\label{eqn:max_perturb}
  \max_{\|\Delta \x\|_p\leq\epsilon} \mathcal{J}(\x+\Delta\x,\theta,\y),
\end{equation}
\noindent where $\epsilon$ constrains the perturbation magnitude, and $\mathcal{J}$ can be either
the loss function ($\mathcal{L}$) or more generally another kernel~\cite{carlini2017towards}.
In the case of DeepFool (DF) attack, this
problem is solved by approximating the decision boundary by a linear affine transformation.
For such a decision boundary, the perturbation magnitude could be analytically computed by
just evaluating the gradient at the current point. However, for neural networks this approximation
could be very inaccurate, that is it could lead to over/under-estimation of the perturbation
along a sub-optimal direction. The smallest direction would be orthogonal to the decision boundary, and this cannot
be computed by a simple affine transformation, as the decision boundary is non-linear (see Fig.~\ref{f:decision_boundary} for illustration).
This is obvious for non-linear activation functions, but even in the case of ReLU, the model behaves like a piece-wise linear function (before the softmax layer, and 
actually non-linear afterwards). 
This approach cannot correctly find the orthogonal direction, even if we ignore the non-linearity of the softmax layer.

To address this limitation, we instead use TR methods, which are well-known for solving non-convex optimization problems~\cite{steihaug1983conjugate}.
The problem of finding adversarial perturbation using TR is defined as follows:

\begin{equation}\label{eqn:tr_prob_adv}
\max_{\|\Delta\x^j\|_p\leq\eps^j} m^j(\Delta\x^j) = 
\langle\Delta\x^j, \g^j_{t,i}\rangle + \frac12\langle\Delta\x^j, \H^j_{t,i}\Delta\x^j\rangle,
\end{equation}

\noindent 
where $\eps^j$ is the TR radius at $j^{th}$ iteration,
$m^j$ is the approximation of the kernel function of $f(\x^{j-1})=(\z_t^{j-1}-\z^{j-1}_i)$ with
$\g^j_{t,i}$ and $\H^j_{t,i}$ denoting the corresponding gradient and Hessian,
and $\x^j = \x + \sum_{i=1}^{j-1}\Delta\x^i$. 
The main idea of the TR method is to iteratively select the trusted radius $\eps^j$ to find the adversarial perturbation within this region such that the probability of an incorrect class becomes maximum.
TR adjusts this radius
by measuring the approximation of the local model
$m^j(\s^j)$ to the actual function value $f(\x^{j+1})-f(\x^j)$.
In particular, we increase the trusted radius if
the approximation of the function is accurate (measured by $\frac{f(\x^{j+1})-f(\x^j)}{m^j(\s^j)}>\sigma_1$ with 
a typical value of $\sigma_1=0.9$). 
In such a case, the trusted radius is increased for the next iterations
 by a factor of $\eta>1$
($\eps^{j+1}=\eta\eps^t$). However, when the local model $m^j(\s^j)$ is a
poor approximation of $f(\x^{j+1})-f(\x^j)$, i.e.,
$\frac{f(\x^{j+1})-f(\x^j)}{m^j(\s^j)}<\sigma_2$ (with a typical $\sigma_2=0.5$), we decrease the trusted radius for the next iteration
$\eps^{j+1}=\eps^t/\eta$.
Otherwise, we keep the same $\eps^j$ for $\eps^{j+1}$.
Typically, a threshold is also used for lower and upper bounds of $\eps^j$.
Using this approach, the TR attack can iteratively find an adversarial perturbation to fool the network. 
See Alg.~\ref{alg:1} for details.

Note that for cases where all the activations of the DNN are ReLU, the Hessian is
zero almost everywhere~\cite[Theorem 1]{yao2018hessian}, and we actually do not need the Hessian.
This means the landscape of $\z_t-\z_i$ is piece-wise linear, i.e., we could omit $\H^j_{t,i}$ in \eqref{eqn:tr_prob_adv}.
However, for non-linear activation functions, we need to keep the Hessian term (since when the NN has smooth activation functions, the Hessian is not zero).
For these cases, the problem of finding the adversarial perturbation becomes a Quadratic Constrained Quadratic Programming (QCQP) problem.
It is quadratic constraint due to the fact that the norm of the perturbation is limited by the TR radius, $\eta^j$, and
the quadratic programming arises from the non-zero Hessian term. We use Lanczos algorithm to solve the QCQP problem. In
this approach, the solution is iteratively found in a Krylov subspace formed by the Hessian operator.

\begin{table*}[!htbp]
\caption{Average perturbations / worst case perturbations 
are reported of different models on Cifar-10. 
Lower values are better.
The first set of rows show $L_2$ attack and the second shows $L_\infty$ attack. 
}
\label{tab:cifar}
\centering
\small
\setlength\tabcolsep{4.1pt}
\begin{tabular}{lcc|c|c|ccccccc} \toprule
             &                      &DeepFool        & CW               &TR Non-Adap              &TR Adap \\ 
             \cmidrule{3-6}
     Model           & Accuracy       &  {$\rho_2$}           &  {$\rho_2$}        &  {$\rho_2$}  &  {$\rho_2$}           \\          
    \midrule
\Gc  AlexLike        & 85.78     & 1.67\% / 11.5\%   & 1.47\% / 9.70\%        & 1.49\% / 9.13\%     & 1.49\% / 9.09\%       \\
\Ga  AlexLike-S      & 86.53     & 1.74\% / 11.0\%   & 1.57\% / 8.59\%        & 1.57\% / 9.48\%     & 1.57\% / 9.46\%     \\ 
\Gc  ResNet          & 92.10     & 0.80\% / 5.60\%   & 0.62\% / 3.12\%        & 0.66\% / 3.97\%     & 0.66\% / 3.96\%     \\
\Ga  WResNet         & 94.77     & 0.89\% / 5.79\%   & 0.66\% / 4.51\%        & 0.73\% / 4.43\%     & 0.72\% / 4.34\%     \\     
\midrule
            &                      &DeepFool            & FGSM                         &TR Non-Adap              &TR Adap \\ 
            \cmidrule{3-6}
     Model           & Accuracy       &  {$\rho_\infty$}  &  {$\rho_\infty$}         &  {$\rho_\infty$}   &  {$\rho_\infty$}       \\          
    \midrule
\Gc  AlexLike        & 85.78          & 1.15\% / 6.85\%   & 1.40\% / 16.44\%       & 1.05\% / 5.45\%     & 1.03\% / 5.45\%           \\
\Ga  AlexLike-S      & 86.53          & 1.18\% / 6.01\%   & 1.45\% / 14.88\%       & 1.09\% / 4.76\%     & 1.07\% / 4.73\%           \\
\Gc  ResNet          & 92.10          & 0.60\% / 3.98\%   & 0.85\% / 4.35\%        & 0.56\% / 3.18\%     & 0.50\% / 3.35\%           \\
\Ga  WResNet         & 94.77          & 0.66\% / 3.34\%   & 0.85\% / 3.30\%        & 0.56\% / 2.67\%     & 0.54\% / 2.69\%           \\        
     \bottomrule 
\end{tabular}
\end{table*}

\section{Performance of the Method}
\label{sec:results}

To test the efficacy of the TR attack method and to compare its performance with other approaches, we perform multiple experiments using different models on Cifar-10~\cite{krizhevsky2009learning} and ImageNet~\cite{deng2009imagenet} datasets.
In particular, we compare to DeepFool~\cite{moosavi2016deepfool}, iterative FGSM~\cite{fgsm,fgsm10}, and the Carlini-Wagner (CW) attack~\cite{carlini2017towards}.

As mentioned above, the original TR method adaptively selects the perturbation magnitude.
Here, to test how effective the adaptive method performs, we also experiment with a case where we set the TR radius to be a fixed small value and compare the results with the original adaptive version.
We refer to the fixed radius version as "TR Non-Adap" and the adaptive version as "TR Adap". 
Furthermore, the metric that we use for performance of the attack is the relative perturbation, defined as follows:
\begin{equation}\label{eqn:rel_per}
\rho_p = \frac{\|\Delta\x\|_p}{\|\x\|_p},
\end{equation}
where $\Delta\x$ is the perturbation needed to fool the testing example. 
The perturbation is chosen such that the accuracy of the model is reduced
to less than $0.1\%$. We report both the average perturbation as well as the highest
perturbation required to fool a testing image. To clarify this, the highest
perturbation is computed after all of testing images (50K in ImageNet and 10K in Cifar-10) and then finding the
the highest perturbation magnitude that was needed to fool a correctly classified example. We refer to this case
as \textit{worst case} perturbation. Ideally we would like this worst case
perturbation to be bounded and close to the average cases.

\begin{table}[!htbp]
\caption{ Average perturbations / worst case perturbations
are reported of different models on Cifar-10 for hardest class attack. 
Lower values are better.
The first set of rows show $L_2$ attack and the second shows $L_\infty$ attack. 
}
\label{tab:cifar_worst}
\small
\setlength\tabcolsep{2.85pt} 
\centering
\begin{tabular}{lc|c|ccccccccccc} \toprule
                   &DeepFool   &TR Non-Adap      &TR Adap\\ 
\midrule
     Model           &  {$\rho_2$}       &  {$\rho_2$}   &  {$\rho_2$}         \\          
    \midrule
\Gc  AlexLike            & 4.36\% /18.9\%  & 2.47\% /13.4\%  & 2.47\% /13.4\%         \\
\Ga  AlexLike-S          & 4.70\% /17.7\%  & 2.63\% /14.4\%  & 2.62\% /14.2\%       \\     
\Gc  ResNet              & 1.71\% /8.01\%  & 0.99\% /4.76\%  & 0.99\% /4.90\%         \\
\Ga  WResNet             & 1.80\% /8.74\%  & 1.05\% /6.23\%  & 1.08\% /6.23\%        \\    
     \midrule
     Model            &  {$\rho_\infty$}        &  {$\rho_\infty$}     &  {$\rho_\infty$}           \\          
    \midrule
\Gc  AlexLike     & 2.96\% /12.6\%         & 1.92\% /9.99\%    & 1.86\% /10.0\%           \\
\Ga  AlexLike-S   & 3.12\% /12.2\%         & 1.98\% /8.19\%    & 1.92\% /8.17\%           \\
\Gc  ResNet       & 1.34\% /9.65\%         & 0.77\% /4.70\%    & 0.85\% /5.44\%           \\
\Ga  WResNet      & 1.35\% /6.49\%         & 0.81\% /3.77\%    & 0.89\% /3.90\%      \\  
\bottomrule
\end{tabular}
\end{table}
\begin{table*}[!htbp]
\caption{ Average perturbations / worst case perturbations
are reported of different models on ImageNet.
Lower values are better.
The first set of rows show $L_2$ attack and the second shows $L_\infty$ attack. 
}
\label{tab:imagenet}
\centering
\small
\setlength\tabcolsep{4.1pt} 
\begin{tabular}{lcc|c|c|ccccccccccc} 
\toprule
             &                      &DeepFool        & CW               &TR Non-Adap              &TR Adap \\ 
             \cmidrule{3-6}
     Model           & Accuracy       &  {$\rho_2$}           &  {$\rho_2$}        &  {$\rho_2$}  &  {$\rho_2$}           \\   
\midrule
\Ga  AlexNet          & 56.5   & 0.20\% / 4.1\%  & 0.31\% / 1.8\%  & 0.17\% / 2.5\%     & 0.18\% / 3.3\%     \\
\Gc  VGG16            & 71.6   & 0.14\% / 4.4\%  & 0.16\% / 1.1\%  & 0.12\% / 1.2\%     & 0.12\% / 3.8\%     \\
\Ga  ResNet50         & 76.1   & 0.17\% / 3.0\%  & 0.19\% / 1.1\%  & 0.13\% / 1.5\%     & 0.14\% / 2.3\%     \\
\Gc  DenseNet121      & 74.4   & 0.15\% / 2.5\%  & 0.20\% / 1.5\%  & 0.12\% / 1.3\%     & 0.13\% / 1.7\%     \\
\midrule
            &                      &DeepFool            & FGSM                         &TR Non-Adap              &TR Adap \\ 
            \cmidrule{3-6}
     Model           & Accuracy       &  {$\rho_\infty$}  &  {$\rho_\infty$}         &  {$\rho_\infty$}   &  {$\rho_\infty$}       \\           
    \midrule
\Ga  AlexNet          & 56.5         & 0.14\% / 4.3\%  & 0.16\% / 4.7\%      & 0.13\% / 1.4\%       & 0.13\% / 3.6\%       \\
\Gc  VGG16            & 71.5         & 0.11\% / 4.0\%  & 0.18\% / 5.1\%      & 0.10\% / 1.4\%       & 0.10\% / 3.4\%       \\
\Ga  ResNet50         & 76.1         & 0.13\% / 3.2\%  & 0.18\% / 3.7\%      & 0.11\% / 1.3\%       & 0.11\% / 2.7\%      \\
\Gc  DenseNet121      & 74.4         & 0.11\% / 2.3\%  & 0.15\% / 4.1\%      & 0.10\% / 1.1\%       & 0.10\% / 1.8\%       \\
     \bottomrule    
\end{tabular}
\end{table*}

\subsection{Setup}\label{sec:outline}
We consider multiple different neural networks including variants of
(wide) residual networks~\cite{he2016deep,zagoruyko2016wide}, AlexNet, VGG16~\cite{simonyan2014very}, and DenseNet from~\cite{huang2017densely}.
We also test with custom smaller/shallower convolutional networks such as a simple
CNN~\cite[C1]{yao2018hessian} (refer as AlexLike with ReLU and AlexLike-S with Swish activation). 
To test the second order attack method we run experiments
with AlexNet-S (by replacing all ReLUs with Swish function activation function~\cite{ramachandran2017swish}), along with a
simple MLP ($3072\rightarrow1024\rightarrow512\rightarrow512\rightarrow256\rightarrow10$) with Swish activation function.

By definition, an adversarial attack is considered successful if it is able
to change the classification of the input image.
Here we perform two types of attacks. The first one is where we
compute the smallest perturbation needed to change the target label. We refer to this as \textit{best class} attack.
This means we attack the class with:
    \[
     \argmin_{j} \frac{z_t-z_j}{\|\nabla_\x (z_t-z_j)\|}.
    \]
Intuitively, this corresponds to perturbing the input to cross the closest decision boundary (\fref{f:decision_boundary}). On the other hand, we also 
consider perturbing the input to the class whose decision boundary is farthest away:
    \[
     \argmax_{j} \frac{z_t-z_j}{\|\nabla_\x (z_t-z_j)\|}.
    \]
\noindent Furthermore, we report two perturbation metrics of
\textit{average perturbation}, computed as:
    \[
    \rho_p = \frac{1}{N}\sum_{i=1}^N\frac{\|\Delta\x_i\|_p}{\|\x_i\|_p},
    \]
\noindent along with worst perturbation, computed as:
    \[
    \rho_p = \max\large\{\frac{\|\Delta\x_i\|_p}{\|\x_i\|_p}\large\}_{i=1}^N.
    \]

\noindent 
For comparison, we also consider the following attack methods:

\begin{itemize}
    \item Iterative FGSM from~\cite{fgsm, fgsm10}, where the following formula is used to compute adversarial perturbation, after which the perturbation is clipped
    in range $(\min(\x), \max(x))$:
    \[
        \x^{j+1} = \x^j + \epsilon~sign(\nabla_\x \mathcal{L}(\x^j, \theta, \y)),
    \]
    \item DeepFool (DF) from~\cite{moosavi2016deepfool}. We follow the same implementation as~\cite{moosavi2016deepfool}. For the hardest class test, the
    target class is set as same as our TR method.
    \item CW attack from~\cite{carlini2017towards}. We use the open source code from~\cite{rauber2017foolbox}\footnote{https://github.com/bethgelab/foolbox}.
    
\end{itemize}

Finally, we measure the time to fool an input image by averaging the
attack time over all the testing examples. The measurements are performed
on a Titan Xp GPU with an Intel E5-2640 CPU.

\begin{table}[!htbp]
\caption{ Average perturbations / worst case perturbations 
are reported of different models on ImageNet for hardest class attack (on the top 100 prediction classes).
Lower values are better.
The first set of rows show $L_2$ attack and the second shows $L_\infty$ attack.
}
\label{tab:imagenet_worst}
\centering
\small
\setlength\tabcolsep{4.1pt} 
\begin{tabular}{lc|c|cccccccccccc} 
\toprule
                    &DeepFool   &TR Non-Adap &TR Adap  \\ 
\midrule
     Model                &  {$\rho_2$}        &  {$\rho_2$}        &  {$\rho_2$}                             \\      
\midrule
\Gc  AlexNet            & 0.74\% /8.7\%   & 0.39\% /5.0\%     & 0.39\% /5.0\%     \\
\Ga  VGG16              & 0.45\% /5.4\%   & 0.27\% /3.6\%     & 0.27\% /3.8\%     \\
\Gc  ResNet50           & 0.52\% /5.8\%   & 0.31\% /4.2\%     & 0.31\% /4.2\%     \\
\Ga  DenseNet           & 0.48\% /5.7\%   & 0.29\% /3.8\%     & 0.29\% /3.8\%     \\
\midrule
     Model              &  {$\rho_\infty$}    &  {$\rho_\infty$}          &  {$\rho_\infty$}             \\          
    \midrule
\Gc  AlexNet                & 0.53\% /9.9\%         & 0.31\% /7.5\%      & 0.33\% /9.1\%       \\
\Ga  VGG16                  & 0.36\% /11.6\%        & 0.25\% /5.1\%      & 0.26\% /6.8\%      \\
\Gc  ResNet50               & 0.43\% /6.6\%         & 0.28\% /3.7\%      & 0.30\% /4.6\%     \\
\Ga  DenseNet               & 0.38\% /6.4\%         & 0.24\% /4.5\%      & 0.27\% /5.7\%      \\
     \bottomrule    
\end{tabular}
\end{table}

\begin{table*}[!htbp]
\caption{Second order and first order comparison on MLP and AlexNet with Swish activation 
function on Cifar-10. The corresponding baseline accuracy without adversarial perturbation is 62.4\% and 76.6\%, respectively.
As expected, the second order TR attack achieves better results as compared to first-order with fixed iterations. However, the second-order
attack is significantly more expensive, due to the overhead of solving QCQP problem.}
\label{tab:tr_second}
\centering
\small 
\setlength\tabcolsep{4.1pt} 
\begin{tabular}{llcccccccccccc} \toprule
     Iter       &     & 1       & 2       & 3          & 4   & 5     & 6       &7         & 8     & 9    & 10\\
    \midrule
\Gc  MLP        &  TR First      & 47.63   & 33.7     & 22.24   & 13.76 & 8.13   & 4.59    & 2.41     & 1.31  & 0.63 & 0.27\\
\Ga  MLP        &   TR Second     & 47.84   & 33.37    & 21.49   & 13.3  & 7.39   & 4.16    & 2.17     & 1.09  & 0.49 & 0.20\\
\midrule
\Gc  AlexNet    &  TR First      & 51.51   & 28.17   & 12.45     &5.53   & 2.61  & 1.33    & 0.82   & 0.66   & 0.51  & 0.46\\
\Ga  AlexNet    &  TR Second     & 50.96   & 26.97   & 10.73     &4.11   & 1.79  & 0.91    & 0.67   & 0.54   & 0.47  & 0.44\\
     \bottomrule 
\end{tabular}
\end{table*}

\subsection{Cifar-10}\label{sec:cifar10}
We first compare different attacks of various neural network models on Cifar-10 dataset, as reported in~\tref{tab:cifar}.
Here, we compute the average and worst case perturbation for 
best class attack. For $L_2$ attack, we can see that TR Non-Adap can achieve comparable 
perturbation as CW, with both TR and CW requiring smaller perturbation
than DeepFool.
An important advantage of the TR attack is its speed, as compared to
CW attack, which is illustrated in~\fref{f:efficient_cifar} (please see appendix). Here we plot the time spent to
fool one input image versus average perturbation for all $L_2$ attack methods on 
different models. It can be clearly seen that, with similar perturbations, the time to get the adversarial examples is: TR $<$ CW.
Note that DeepFool is also very fast but requires much larger perturbations than TR attack and CW.
Also note that the TR Adap method achieves similar results, with slightly slower
speed and slightly larger perturbation. This is because the adaptive method
has not been tuned any way, whereas for the non adaptive version we manually
tuned $\eps$.
TR Adap does not require tuning, as it automatically adjust  the TR radius.
The slight performance degradation is due to the relaxed $\sigma_1$ and $\sigma_2$ parameters, which could be made more conservative as a trade-off for speed.
But we did not tune these parameters beyond the default, to give a realistic
performance for the non-tuned version.

Another important test is to measure the perturbation needed to fool
the network to the hardest target class. This is important in that flipping a pedestrian to a cyclist may be easier than flipping it to become a car.
In Table~\ref{tab:cifar_worst}, we report the hardest class attack on Cifar-10. Note
that our methods are roughly $1.5$ times better than DeepFool in all cases. 
Particularly, For $L_2$ attack on WResNet, our worst case is $3.9$ times better than
DeepFool in terms of perturbation magnitude.

\subsection{ImageNet Result}\label{sec:imagenet}
We observe similar trends on ImageNet.
We report different attacks on various models 
on ImageNet in Table~\ref{tab:imagenet}. Note that TR and CW require
significantly smaller perturbation for the worst case as compared to DeepFool.
However, TR is significantly faster than CW. The timing results are shown in~\fref{f:efficient}.
For instance in the case of VGG-16, TR attack is $\mathbf{37.5\times}$ faster than CW which is significant.
An  example perturbation with AlexNet is shown in~\fref{f:adv_example1} (for which TR is $\bf 15\btimes$ faster).
As one can see, CW and TR 
perturbations are smaller than DeepFool ($2\times$ in this case), and 
more targeted around the object. For $L_\infty$ methods, our TR Non-Adap and TR Adap
are consistently better than FGSM and DeepFool in both average and worst cases. 
Particularly, for worst cases, TR is roughly two times better than the other methods. 
An example perturbation of DeepFool and TR Non-Adap with $L_\infty$ on VGG16 is shown in~\fref{f:adv_example2}. It can be clearly seen that, TR perturbation is much smaller than DeepFool ($1.9\times$ in this case), and more targeted around the objective.

\subsection{Second order method}\label{sec:second_order}
As mentioned in 
Section~\ref{sec:method}, the ReLU activation function does not require Hessian 
computation. However, for non-linear activation functions including Hessian 
information is beneficial, although it may be very expensive. To test this 
hypothesis, we consider two models with Swish activation function. We fix the 
TR radius (set to be 1 for all cases) of our first and second order 
methods, and gradually increase the iterations. Table~\ref{tab:tr_second} shows the results for MLP and AlexNet models.
It can be seen that second order TR out-performs the first order TR method in all 
iterations. Particularly, for two and three iterations on AlexNet, TRS can drop the model 
accuracy 1.2\% more as compared to the first order TR variant. However, the second
order based model is more expensive than the first order model, mainly due to
the overhead associated with solving the QCQP problem. There is no closed form
solution for this problem because the problem is non-convex and the Hessian can contain negative
spectrum. Developing computationally efficient method for this is an interesting next direction.

\section{Conclusions}\label{sec:conclusion}
We have considered various TR based methods for adversarial attacks on neural networks.
We presented the formulation for the TR method along with results for our first/second-order attacks.
We considered multiple models on Cifar-10 and ImageNet datasets, including variants of residual and densely connected networks. 
Our method requires significantly smaller perturbation (up to $3.9\times$), as compared to DeepFool. 
Furthermore, we achieve similar results (in terms of 
average/worst perturbation magnitude to fool the network), as compared to the CW attack, but with significant speed up of up to $37.5\times$.
For all the models considered, our attack method can bring down the model accuracy to less than $0.1\%$ with relative small perturbation (in $L_2/L_\infty$ norms) of the input image. 
Meanwhile, we also tested the second order TR attack by backpropogating the Hessian information through the neural network, showing that it can find a stronger attack direction, as compared to the first order variant.

\clearpage
{\small
\bibliographystyle{plain}
\bibliography{ref}

\begin{thebibliography}{10}

\bibitem{TRcode}
https://github.com/amirgholami/trattack, November 2018.

\bibitem{athalye2018obfuscated}
Anish Athalye, Nicholas Carlini, and David Wagner.
\newblock Obfuscated gradients give a false sense of security: Circumventing
  defenses to adversarial examples.
\newblock {\em arXiv preprint arXiv:1802.00420}, 2018.

\bibitem{brown2017adversarial}
Tom~B Brown, Dandelion Man{\'e}, Aurko Roy, Mart{\'\i}n Abadi, and Justin
  Gilmer.
\newblock Adversarial patch.
\newblock {\em arXiv preprint arXiv:1712.09665}, 2017.

\bibitem{carlini2017adversarial}
Nicholas Carlini and David Wagner.
\newblock Adversarial examples are not easily detected: Bypassing ten detection
  methods.
\newblock In {\em Proceedings of the 10th ACM Workshop on Artificial
  Intelligence and Security}, pages 3--14. ACM, 2017.

\bibitem{carlini2017towards}
Nicholas Carlini and David Wagner.
\newblock Towards evaluating the robustness of neural networks.
\newblock In {\em 2017 IEEE Symposium on Security and Privacy (SP)}, pages
  39--57. IEEE, 2017.

\bibitem{deng2009imagenet}
Jia Deng, Wei Dong, Richard Socher, Li-Jia Li, Kai Li, and Li~Fei-Fei.
\newblock Imagenet: A large-scale hierarchical image database.
\newblock In {\em Computer Vision and Pattern Recognition, 2009. CVPR 2009.
  IEEE Conference on}, pages 248--255. Ieee, 2009.

\bibitem{fgsm}
Ian~J Goodfellow, Jonathon Shlens, and Christian Szegedy.
\newblock Explaining and harnessing adversarial examples.
\newblock {\em International Conference on Learning Representations
  (arXiv:1412.6572)}, 2015.

\bibitem{he2016deep}
Kaiming He, Xiangyu Zhang, Shaoqing Ren, and Jian Sun.
\newblock Deep residual learning for image recognition.
\newblock In {\em Proceedings of the IEEE conference on computer vision and
  pattern recognition}, pages 770--778, 2016.

\bibitem{huang2017densely}
Gao Huang, Zhuang Liu, Laurens Van Der~Maaten, and Kilian~Q Weinberger.
\newblock Densely connected convolutional networks.
\newblock In {\em CVPR}, volume~1, page~3, 2017.

\bibitem{keskar2016large}
Nitish~Shirish Keskar, Dheevatsa Mudigere, Jorge Nocedal, Mikhail Smelyanskiy,
  and Ping Tak~Peter Tang.
\newblock On large-batch training for deep learning: Generalization gap and
  sharp minima.
\newblock {\em arXiv preprint arXiv:1609.04836}, 2016.

\bibitem{krizhevsky2009learning}
Alex Krizhevsky and Geoffrey Hinton.
\newblock Learning multiple layers of features from tiny images.
\newblock Technical report, Citeseer, 2009.

\bibitem{fgsm10}
Alexey Kurakin, Ian Goodfellow, and Samy Bengio.
\newblock Adversarial examples in the physical world.
\newblock {\em arXiv preprint arXiv:1607.02533}, 2016.

\bibitem{metzen2017detecting}
Jan~Hendrik Metzen, Tim Genewein, Volker Fischer, and Bastian Bischoff.
\newblock On detecting adversarial perturbations.
\newblock {\em arXiv preprint arXiv:1702.04267}, 2017.

\bibitem{moosavi2017universal}
Seyed-Mohsen Moosavi-Dezfooli, Alhussein Fawzi, Omar Fawzi, and Pascal
  Frossard.
\newblock Universal adversarial perturbations.
\newblock In {\em 2017 IEEE Conference on Computer Vision and Pattern
  Recognition (CVPR)}, pages 86--94. IEEE, 2017.

\bibitem{moosavi2016deepfool}
Seyed~Mohsen Moosavi~Dezfooli, Alhussein Fawzi, and Pascal Frossard.
\newblock Deepfool: a simple and accurate method to fool deep neural networks.
\newblock In {\em Proceedings of 2016 IEEE Conference on Computer Vision and
  Pattern Recognition (CVPR)}, number EPFL-CONF-218057, 2016.

\bibitem{NW06}
J.~Nocedal and S.~Wright.
\newblock {\em Numerical Optimization}.
\newblock Springer, New York, 2006.

\bibitem{papernot2016distillation}
Nicolas Papernot, Patrick McDaniel, Xi~Wu, Somesh Jha, and Ananthram Swami.
\newblock Distillation as a defense to adversarial perturbations against deep
  neural networks.
\newblock In {\em 2016 IEEE Symposium on Security and Privacy (SP)}, pages
  582--597. IEEE, 2016.

\bibitem{ramachandran2017swish}
Prajit Ramachandran, Barret Zoph, and Quoc~V Le.
\newblock Swish: a self-gated activation function.
\newblock {\em arXiv preprint arXiv:1710.05941}, 2017.

\bibitem{ramachandran2018searching}
Prajit Ramachandran, Barret Zoph, and Quoc~V Le.
\newblock Searching for activation functions.
\newblock {\em arXiv preprint arXiv:1710.05941}, 2018.

\bibitem{rauber2017foolbox}
Jonas Rauber, Wieland Brendel, and Matthias Bethge.
\newblock Foolbox v0. 8.0: A python toolbox to benchmark the robustness of
  machine learning models.
\newblock {\em arXiv preprint arXiv:1707.04131}, 2017.

\bibitem{sankaranarayanan2017regularizing}
Swami Sankaranarayanan, Arpit Jain, Rama Chellappa, and Ser~Nam Lim.
\newblock Regularizing deep networks using efficient layerwise adversarial
  training.
\newblock {\em arXiv preprint arXiv:1705.07819}, 2017.

\bibitem{robust}
Uri Shaham, Yutaro Yamada, and Sahand Negahban.
\newblock Understanding adversarial training: Increasing local stability of
  supervised models through robust optimization.
\newblock {\em Neurocomputing}, 2018.

\bibitem{simonyan2014very}
Karen Simonyan and Andrew Zisserman.
\newblock Very deep convolutional networks for large-scale image recognition.
\newblock {\em arXiv preprint arXiv:1409.1556}, 2014.

\bibitem{steihaug1983conjugate}
Trond Steihaug.
\newblock The conjugate gradient method and trust regions in large scale
  optimization.
\newblock {\em SIAM Journal on Numerical Analysis}, 20(3):626--637, 1983.

\bibitem{szegedy2013intriguing}
Christian Szegedy, Wojciech Zaremba, Ilya Sutskever, Joan Bruna, Dumitru Erhan,
  Ian Goodfellow, and Rob Fergus.
\newblock Intriguing properties of neural networks.
\newblock {\em arXiv preprint arXiv:1312.6199}, 2013.

\bibitem{tramer2017ensemble}
Florian Tram{\`e}r, Alexey Kurakin, Nicolas Papernot, Ian Goodfellow, Dan
  Boneh, and Patrick McDaniel.
\newblock Ensemble adversarial training: Attacks and defenses.
\newblock {\em arXiv preprint arXiv:1705.07204}, 2017.

\bibitem{yao2018large}
Zhewei Yao, Amir Gholami, Kurt Keutzer, and Michael~W. Mahoney.
\newblock Large batch size training of neural networks with adversarial
  training and second-order information.
\newblock {\em arXiv preprint arXiv:1810.01021}, 2018.

\bibitem{yao2018hessian}
Zhewei Yao, Amir Gholami, Qi~Lei, Kurt Keutzer, and Michael~W Mahoney.
\newblock Hessian-based analysis of large batch training and robustness to
  adversaries.
\newblock {\em Neural Information Processing Systems (NIPS'18)}, 2018.

\bibitem{zagoruyko2016wide}
Sergey Zagoruyko and Nikos Komodakis.
\newblock Wide residual networks.
\newblock {\em arXiv preprint arXiv:1605.07146}, 2016.

\end{thebibliography}
}
 \clearpage
 \onecolumn
 
\section{Appendix}


\subsection{Importance of Speed: Adversarial Training for Large Scale Learning}

As discussed in the introduction, one of the motivations
of using adversarial attacks is that multiple recent studies have shown superior results when adversarial data
is injected during training~\cite{robust,sankaranarayanan2017regularizing,yao2018hessian,yao2018large}.
However, the overhead of computing an adversarial noise needs to be small, specially for large
scale training.
Here, the speed with which the adversarial noise is computed becomes important, so as not
to slow down the training time significantly.
To illustrate the efficiency of TR attacks, we
perform adversarial training using the Adaptive Batch Size Adversarial Training (ABSA) method introduced by~\cite{yao2018hessian,yao2018large} for large scale training.
We test ResNet-18 on Cifar-10 which achieves a baseline accuracy of $\bf83.50\%$ (with a batch size of 128 using the setup as in~\cite{yao2018large}).
If we use the TR $L_\infty$ attack, the final accuracy increases to $\bf87.79\%$ (actually
for the hardest with batch size of 16K).
If we instead use FGSM, the final accuracy only increases to $\bf84.32\%$ (for the same batch size of 16K).
It is important to note that it is computationally infeasible to use CW in this sort of training, as it is about $15\times$  slower than TR/FGSM. 
This result shows that a stronger attack that is fast may be very useful for adversarial training.
We emphasize that the goal of this test is simply to illustrate the importance of a fast and stronger adversarial attack, which could make TR to be a useful tool for adversarial training research.  More thorough testing (left for future work) is required to understand how stronger attacks such as TR could be used in the context of adversarial training.

\subsection{Visual Examples}
Here we provide more visual examples for different neural networks on ImageNet.
We show the original image along with adversarially perturbed ones using the
three methods of DeepFool, CW, and ours (TR). We also provide
two heat maps for the adversarial perturbations. The first one is calibrated
based on DeepFool's maximum perturbation magnitude, and the second one is calibrated for each of the individual method's maximum perturbation. 
This allows a better visualization of the adversarial perturbation of each method.

\begin{figure*}[!b]
\centering
\includegraphics[width=0.5\textwidth]{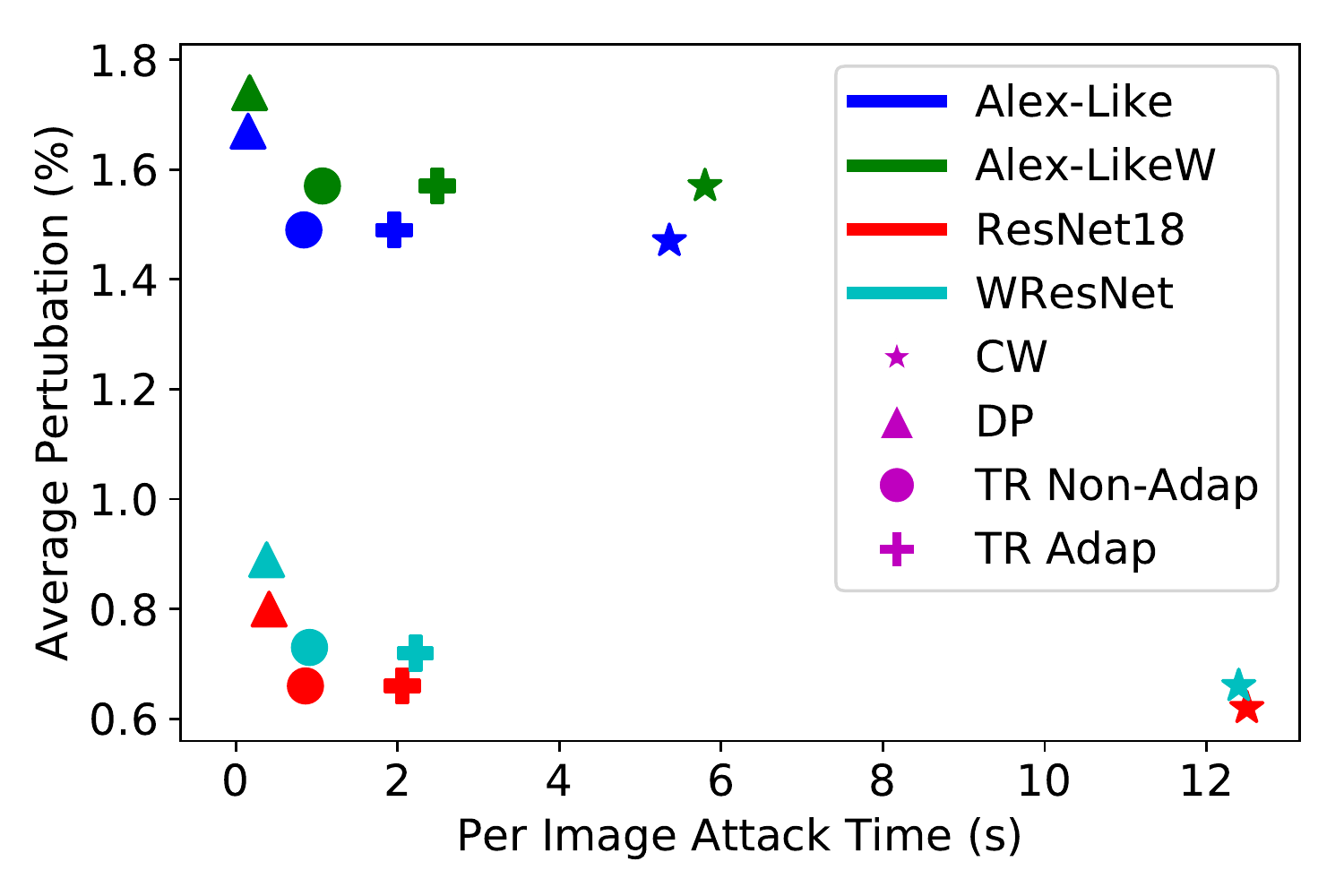}\\
\includegraphics[width=0.5\textwidth]{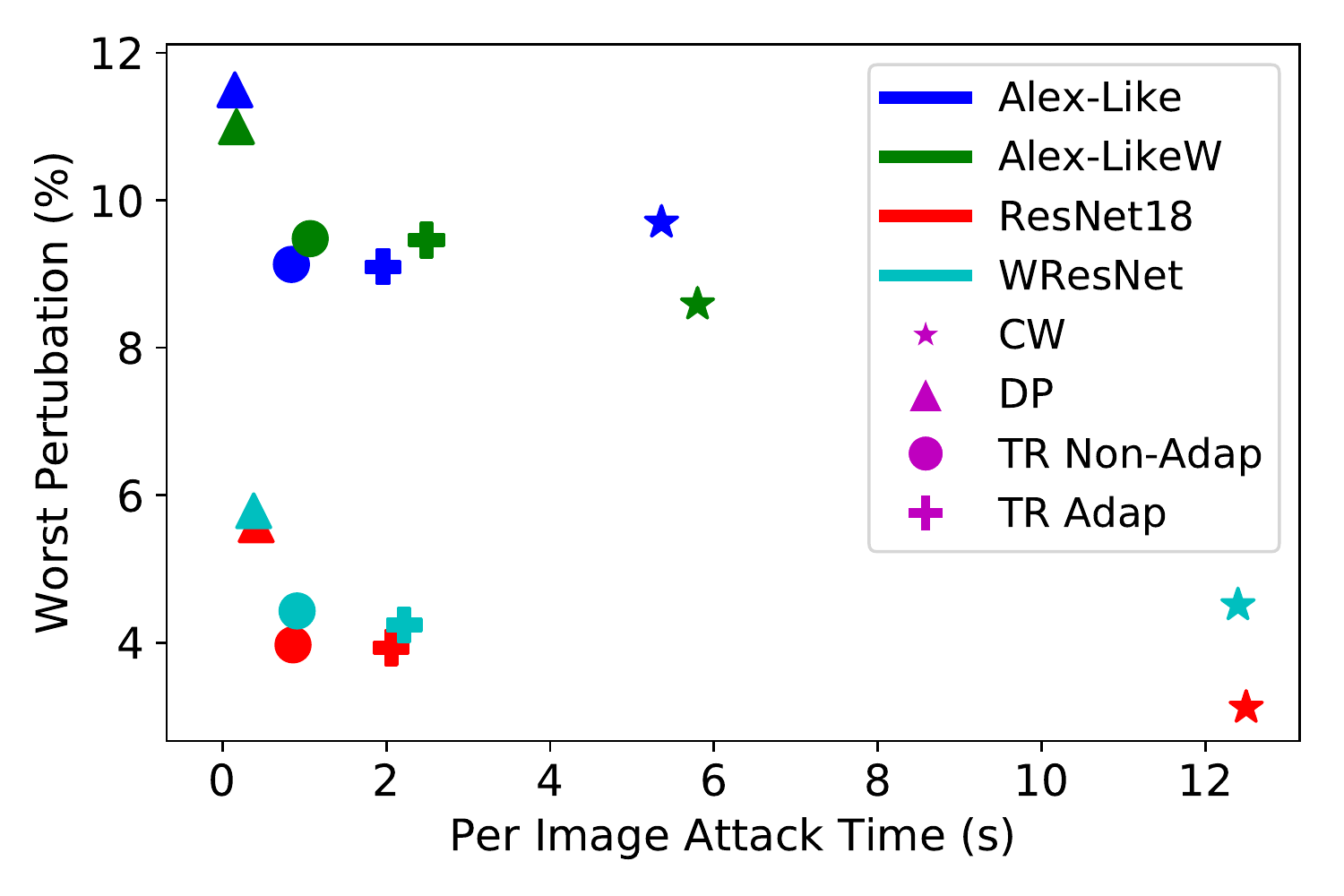} 
\caption{ The figures show various neural networks, the corresponding 
time to compute the adversarial attack (x-axis) and the perturbations a particular 
attack method needs to fool an image (detailed results in~\tref{tab:cifar}).
The results are obtained for the Cifar-10 dataset (for ImageNet results please see~\fref{f:efficient}).
On the 
left we report the average perturbation
and on the right we report the worst perturbation.
In particular,  different colors represent different models, and different markers 
illustrate the different attack methods. 
Notice that, our TR and TR Adap achieve similar perturbations as CW but with 
significantly less time (up to $\bf14.5\btimes$). 
}
\label{f:efficient_cifar}
\end{figure*}

\begin{figure*}[!htbp]
\centering
\includegraphics[width=0.9\textwidth, trim={5cm 1.7cm 2cm 1.7cm},clip]{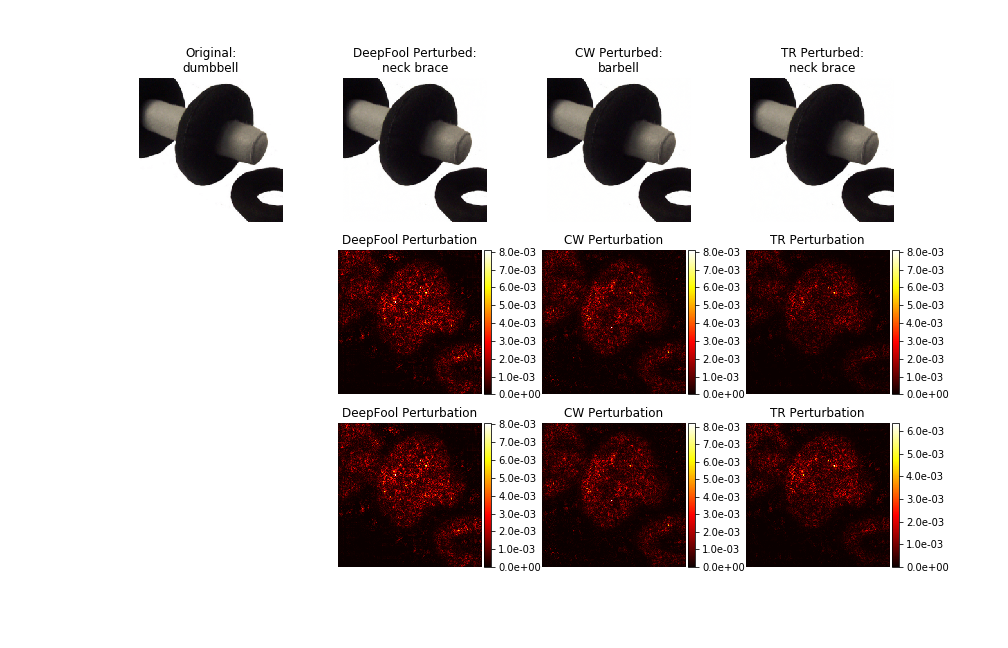}\\
\includegraphics[width=0.9\textwidth, trim={5cm 1.7cm 2cm 1.7cm},clip]{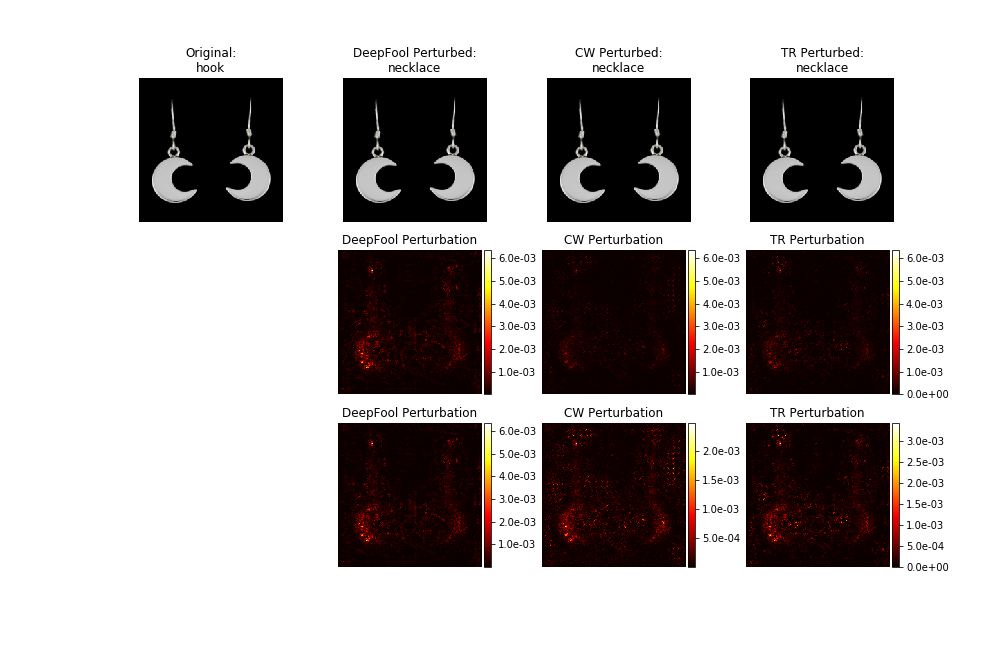} 
\caption{ ResNet-50 adversarial examples; row 1: original and adversarial 
images with ImageNet labels as titles; row 2: perturbation heat map with same
colorbar scale; row 3: perturbations with colorbar scale adjusted to each 
method for better visualization. TR attack obtains similar perturbation as CW, but $\bf26.4\btimes$ faster~\fref{f:efficient}.
}
\label{f:resnet_eg1}
\end{figure*}

\begin{figure*}[!htbp]
\centering
\includegraphics[width=0.9\textwidth, trim={5cm 1.7cm 2cm 1.7cm},clip]{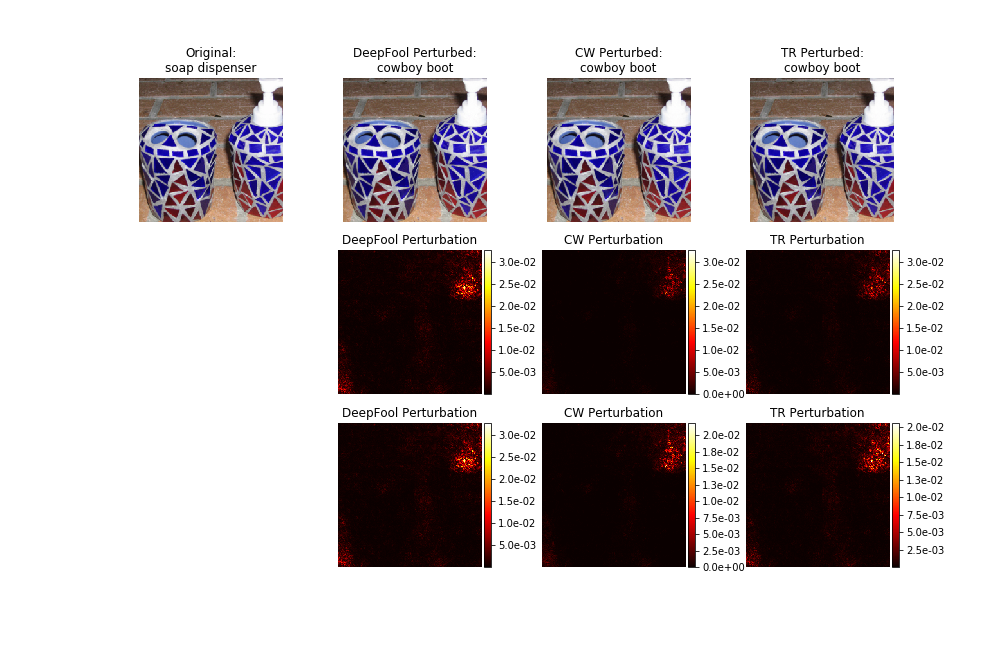}\\
\includegraphics[width=0.9\textwidth, trim={5cm 1.7cm 2cm 1.7cm},clip]{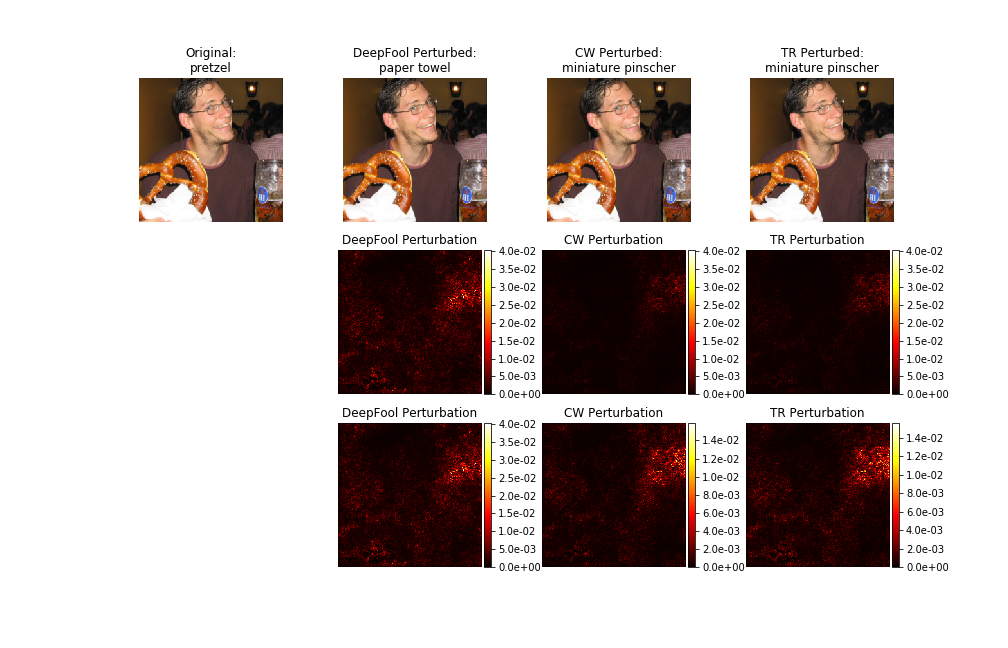} 
\caption{ ResNet-50 adversarial examples; row 1: original and adversarial 
images with ImageNet labels as titles; row 2: perturbation heat map with same
colorbar scale; row 3: perturbations with colorbar scale adjusted to each 
method for better visualization. TR attack obtains similar perturbation as CW, but $\bf26.4\btimes$ faster~\fref{f:efficient}.
}
\label{f:resnet_eg2}
\end{figure*}

\begin{figure*}[!htbp]
\centering
\includegraphics[width=0.9\textwidth, trim={5cm 1.7cm 2cm 1.7cm},clip]{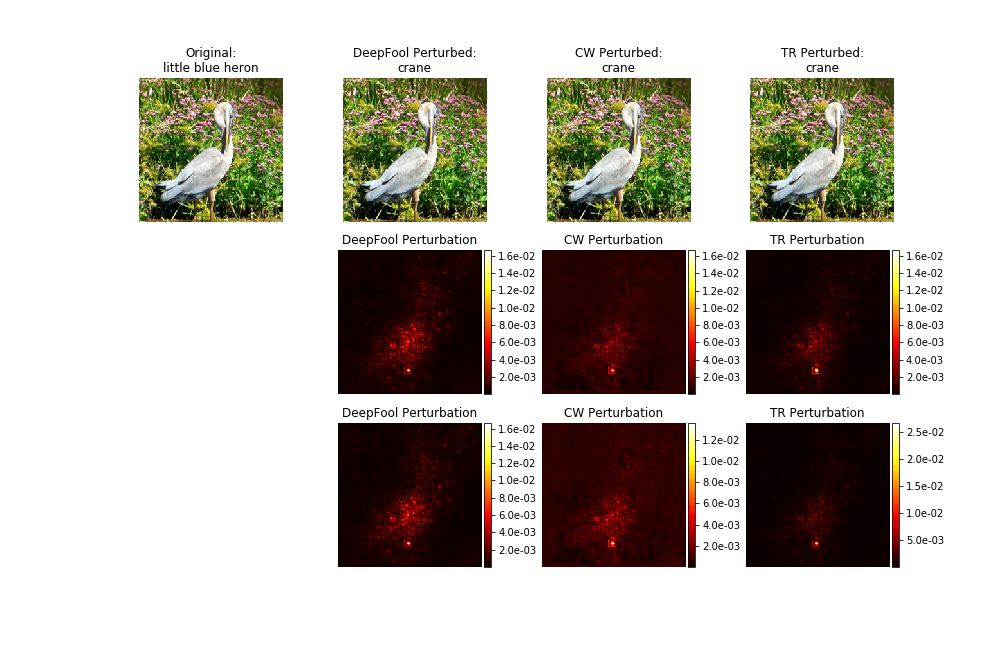}\\
\includegraphics[width=0.9\textwidth, trim={5cm 1.7cm 2cm 1.7cm},clip]{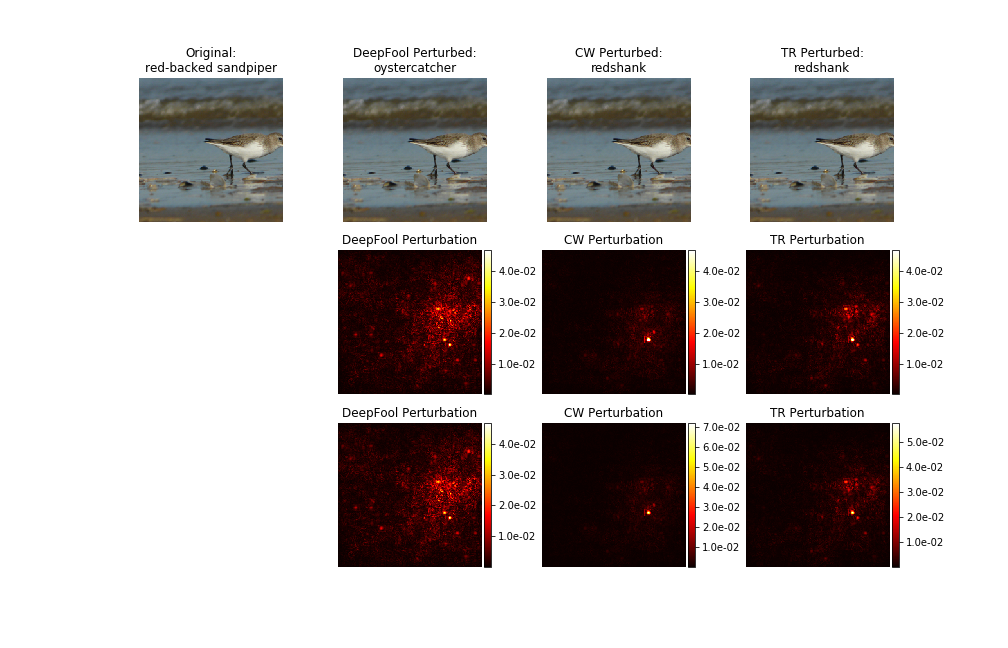} 
\caption{AlexNet adversarial examples; row 1: original and adversarial 
images with ImageNet labels as titles; row 2: perturbation heat map with same
colorbar scale; row 3: perturbations with colorbar scale adjusted to each 
method for better visualization.
TR attack obtains similar perturbation as CW, but $\bf15\btimes$ faster~\fref{f:efficient}.
}
\label{f:alex_eg1}
\end{figure*}

\begin{figure*}[!htbp]
\centering
\includegraphics[width=0.9\textwidth, trim={5cm 1.7cm 2cm 1.7cm},clip]{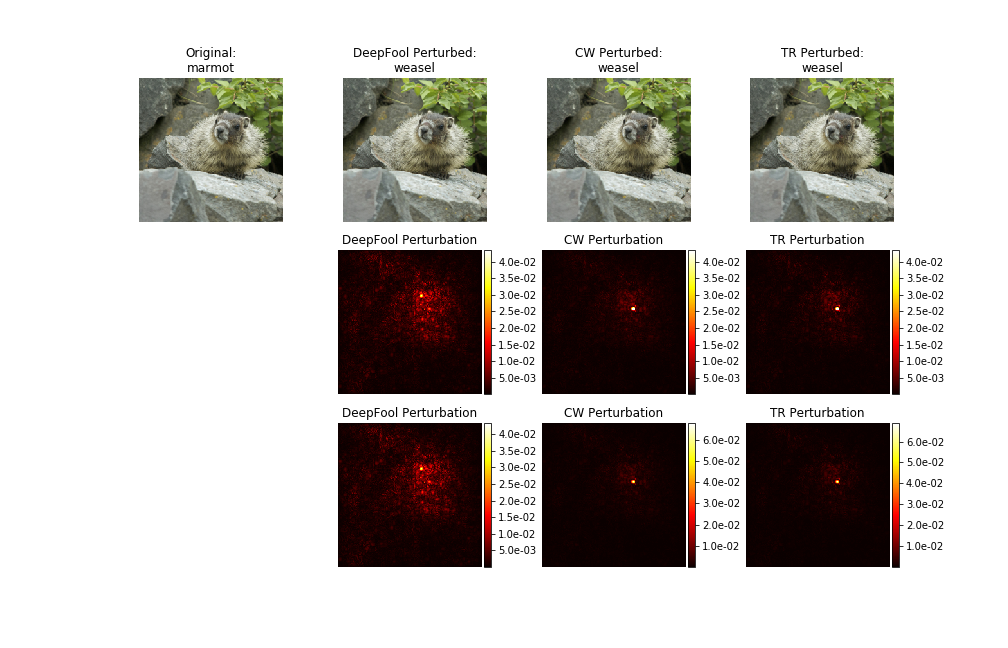}\\
\includegraphics[width=0.9\textwidth, trim={5cm 1.7cm 2cm 1.7cm},clip]{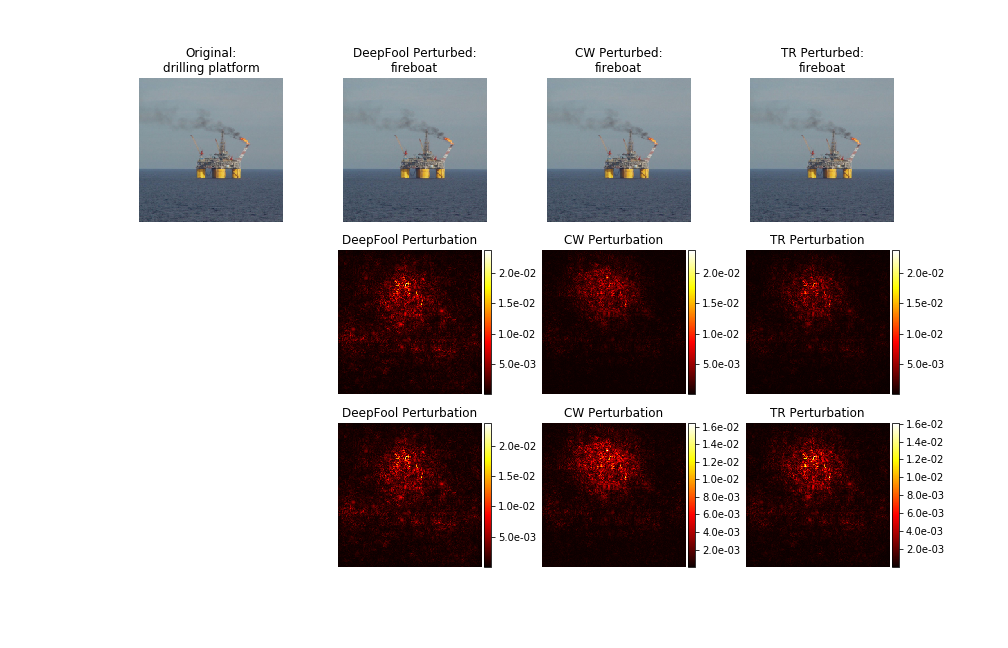} 
\caption{ AlexNet adversarial examples; row 1: original and adversarial 
images with ImageNet labels as titles; row 2: perturbation heat map with same
colorbar scale; row 3: perturbations with colorbar scale adjusted to each 
method for better visualization. TR attack obtains similar perturbation as CW, but $\bf15\btimes$ faster~\fref{f:efficient}.
}
\label{f:alex_eg2}
\end{figure*}

\begin{figure*}[!htbp]
\centering
\includegraphics[width=0.9\textwidth, trim={5cm 1.7cm 2cm 1.7cm},clip]{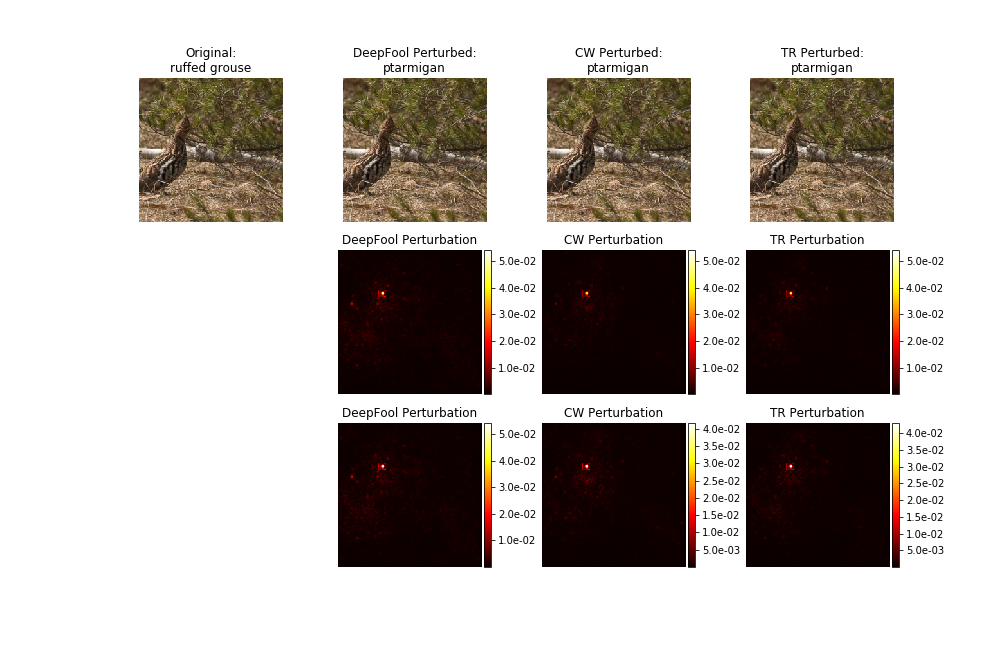}
\caption{ AlexNet adversarial examples; row 1: original and adversarial 
images with ImageNet labels as titles; row 2: perturbation heat map with same
colorbar scale; row 3: perturbations with colorbar scale adjusted to each 
method for better visualization. TR attack obtains similar perturbation as CW, but $15\times$ faster~\fref{f:efficient}.
}
\label{f:alex_eg3}
\end{figure*}



\begin{figure*}[!htbp]
\centering
\includegraphics[width=0.9\textwidth, trim={5cm 1.7cm 2cm 1.7cm},clip]{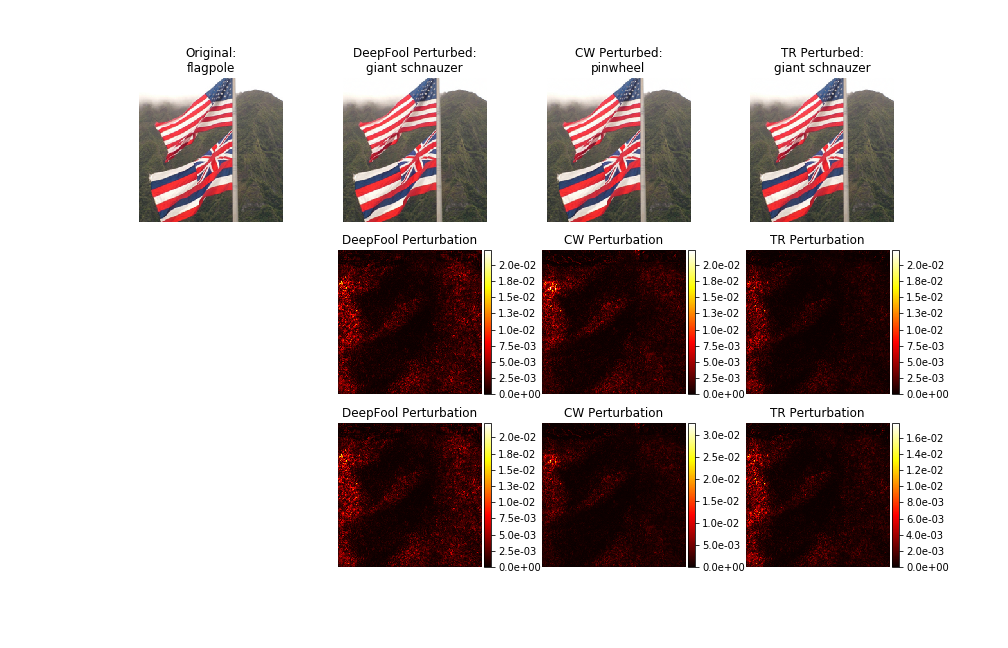}\\
\includegraphics[width=0.9\textwidth, trim={5cm 1.7cm 2cm 1.7cm},clip]{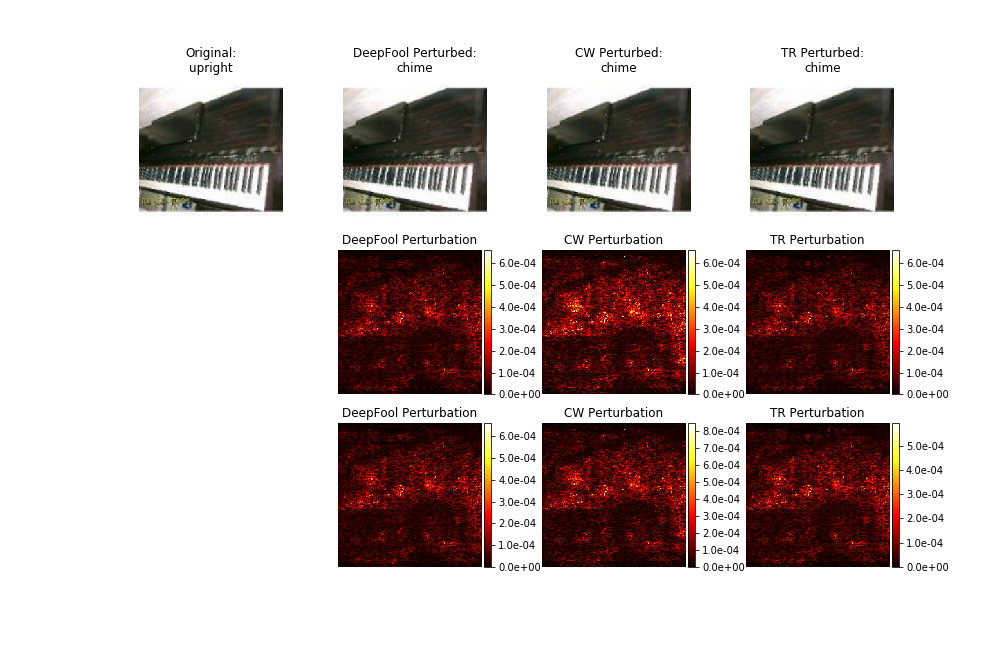} 
\caption{ DenseNet-121 adversarial examples; row 1: original and adversarial 
images with ImageNet labels as titles; row 2: perturbation heat map with same
colorbar scale; row 3: perturbations with colorbar scale adjusted to each 
method for better visualization.  TR attack obtains similar perturbation as CW, but $\bf26.8\btimes$ faster~\fref{f:efficient}.
}
\label{f:densenet_eg1}
\end{figure*}

\begin{figure*}[!htbp]
\centering
\includegraphics[width=0.9\textwidth, trim={5cm 1.7cm 2cm 1.7cm},clip]{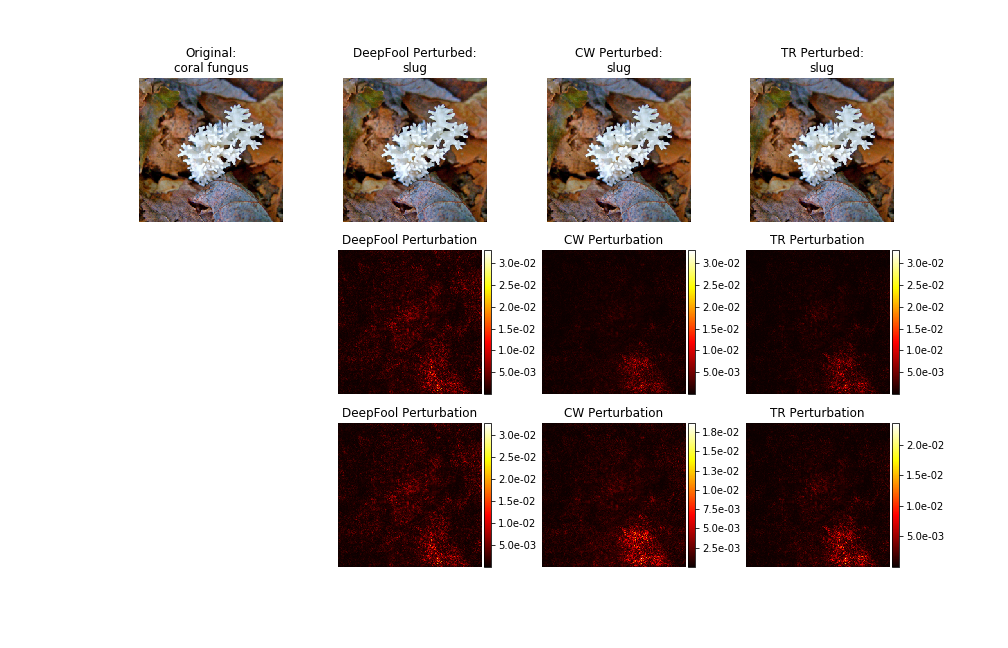}
\caption{ DenseNet-121 adversarial examples; row 1: original and adversarial 
images with ImageNet labels as titles; row 2: perturbation heat map with same
colorbar scale; row 3: perturbations with colorbar scale adjusted to each 
method for better visualization.  TR attack obtains similar perturbation as CW, but $\bf 26.8\btimes$ faster~\fref{f:efficient}.
}
\label{f:densenet_eg2}
\end{figure*}


\end{document}